%% file: main.tex
\documentclass{article}
 \pdfoutput=1

\usepackage[preprint, nonatbib]{neurips_2020}

\usepackage{enumitem}
\usepackage{todonotes}
\usepackage[utf8]{inputenc} 
\usepackage[T1]{fontenc}    
\usepackage{hyperref}       
\usepackage{url}            
\usepackage{booktabs}       
\usepackage{amsfonts}       
\usepackage{nicefrac}       
\usepackage{microtype}      
\usepackage{amsmath}
\usepackage{xcolor}
\usepackage{float}
\usepackage{amssymb}
\usepackage{amsthm}
\usepackage{subfigure}
\usepackage{graphicx}
\usepackage{verbatim}
\usepackage{multirow}
\usepackage{multicol}

\usepackage{titlesec}
\titlespacing\section{0pt}{0pt plus 0pt minus 0pt}{0pt plus 0pt minus 0pt}
\titlespacing\subsection{0pt}{0pt plus 0pt minus 0pt}{0pt plus 0pt minus 0pt}
\titlespacing\subsubsection{0pt}{0pt plus 0pt minus 0pt}{0pt plus 0pt minus 0pt}

\usepackage[skip=0pt]{caption}
\usepackage[ruled]{algorithm2e}
\usepackage{amsmath, amsthm, thm-restate}
\usepackage{chngcntr}

\newtheoremstyle{slanted}
  {0.9\topsep}
  {0.9\topsep}
  {\it}
  {}
  {\bfseries}
  {.}
  {0.5em}
  {}

\theoremstyle{slanted}
\newtheorem{theorem}{Theorem}

\newcommand\blfootnote[1]{%
  \begingroup
  \renewcommand\thefootnote{}\footnote{#1}%
  \addtocounter{footnote}{-1}%
  \endgroup
}

\title{Masked Language Modeling for Proteins via Linearly Scalable Long-Context Transformers}

\author{%
  Krzysztof Choromanski$^{*1}$, Valerii Likhosherstov$^{*2}$, David Dohan$^{*1}$, Xingyou Song$^{*1}$ \\ 
  \textbf{Andreea Gane$^{*1}$,  Tamas Sarlos$^{*1}$,  Peter Hawkins$^{*1}$, Jared Davis$^{*3}$} \\
  \textbf{David Belanger$^{1}$, Lucy Colwell$^{1,2}$, Adrian Weller$^{2,4}$} \\
  $^1$Google $^2$University of Cambridge $^3$DeepMind $^4$Alan Turing Institute \\
}

\begin{document}

\maketitle
\vspace{-0.5cm}
\begin{abstract}
\vspace{-0.3cm}
Transformer models have achieved state-of-the-art results across a diverse range of domains. However, concern over the cost of training the attention mechanism to learn complex dependencies between distant inputs continues to grow. In response, solutions that exploit the structure and sparsity of the learned attention matrix have blossomed. However, real-world applications that involve long sequences, such as biological sequence analysis, may fall short of meeting these assumptions, precluding exploration of these models. To address this challenge, we present a new Transformer architecture, $\mathrm{Performer}$, based on \textit{Fast Attention Via Orthogonal Random features} (FAVOR). Our mechanism scales linearly rather than quadratically in the number of tokens in the sequence, is characterized by sub-quadratic space complexity and does not incorporate any sparsity pattern priors. Furthermore, it provides strong theoretical guarantees: unbiased estimation of the attention matrix and uniform convergence. It is also backwards-compatible with pre-trained regular Transformers. We demonstrate its effectiveness on the challenging task of protein sequence modeling and provide detailed theoretical analysis. 

\end{abstract}
\blfootnote{$^\ast$Equal contribution.}
\footnotetext[1]{Correspondence to \texttt{\{kchoro,lcolwell\}@google.com}.}


\blfootnote{Code for Transformer models on protein data can be found in \textcolor{blue}{\url{github.com/google-research/google-research/tree/master/protein_lm}} and Performer code can be found in \textcolor{blue}{\url{github.com/google-research/google-research/tree/master/performer}}.}
\vspace{-1cm}
\input{intro_related_work}

\input{algorithm}

\input{theory}

\input{experiments}

\vspace{1mm}
\input{conclusion}

\input{broader_impact}

\input{acknowledgements}
\newpage
\bibliographystyle{abbrv}
\bibliography{performers}

\newpage

\input{appendix}

\end{document}

%% file: intro_related_work.tex
\section{Introduction and related work}
\label{sec:intro_related_work}

Transformers \cite{transformer, universal_t} are powerful neural network architectures that have become SOTA in several areas of machine learning including Natural Language Processing (NLP) (e.g. speech recognition \cite{luo}), Neural Machine Translation (NMT) \cite{nmt}, document generation/summarization, time series prediction, generative modeling (e.g. image generation \cite{parmar}), music generation \cite{simon}, and analysis of biological sequences \cite{rives, progen, ingraham2019generative, elnaggar2019end, du2020energy}. Transformers rely on a trainable \textit{attention} mechanism that identifies complex dependencies between the elements of each input sequence (e.g. amino acids within a protein). Unfortunately, a standard Transformer scales quadratically with the number of tokens $L$ in the input sequence, which is prohibitively expensive for large $L$. Several solutions have been proposed to address this issue \cite{longformer, conformer, imputer, sparsetr, image_transformer}. Most approaches restrict the attention mechanism to attend to local neighborhoods \cite{parmar} or incorporate structural priors on attention such as sparsity \cite{sparsetr}, pooling-based compression \cite{compr} clustering/binning/convolution techniques (e.g. \cite{routing_t} which applies $k$-means clustering to learn dynamic sparse attention regions, or \cite{reformer}, where locality sensitive hashing is used to group together tokens of similar embeddings), sliding windows \cite{longformer}, or truncated targeting \cite{chelba}. Thus these approaches do not aim to approximate regular attention, but rather propose simpler and more tractable attention mechanisms, often by incorporating additional constraints (e.g. identical query and key sets as in \cite{reformer}), or by trading regular attention with sparse attention using more layers \cite{sparsetr}. Unfortunately, there is a lack of rigorous guarantees for the representation power produced by such methods, and sometimes the validity of sparsity patterns can only be verified empirically through trial and error by constructing special GPU operations (e.g. either writing C++ CUDA kernels \cite{sparsetr} or using TVMs \cite{longformer}). Other techniques 
which aim to improve the 
time complexity of Transformers include reversible residual layers allowing for one-time activation storage in training \cite{reformer} and shared attention weights \cite{shared_weights}. These constraints may impede application 
to problems that involve long sequences, where approximations of the attention mechanism are not sufficient. Approximations based on truncated back-propagation \cite{transformerxl} are also unable to capture long-distance correlations since the gradients are only propagated inside a localized window.

Recent work has demonstrated that Transformers fit to the amino acid sequences of single proteins learn to accurately predict information about protein structure and function, and can generate new sequences with specific properties \cite{rives, elnaggar2019end, progen}. Approaches that encode 3D protein structural data via Transformer-based models demonstrate improved performance, despite the restriction of attention to the local structural neighborhoods of each node \cite{du2020energy, ingraham2019generative}. These models provide initial promise for protein design applications, but their applicability beyond the design of single proteins is limited because they truncate sequences to 512 or 1024 amino acids. The ability to scale to longer sequences without imposing sparsity constraints would enable the use of Transformers to jointly model multiple concatenated protein sequences and the interactions between them. This follows recent works employing simpler statistical models that predict protein quaternary structure, protein-protein interactions and protein interaction networks from evolutionary sequence data~\cite{weigt2009identification, hopf2012three, ovchinnikov2014robust, bitbol2016inferring, cong2019protein}. 

In response, we present a new Transformer architecture, $\mathrm{Performer}$, based on \textit{Fast Attention Via Orthogonal Random features} (FAVOR). Our proposed mechanism has several advantageous properties: it scales linearly rather than quadratically in the number of tokens in the sequence (important for analysis involving multiple protein molecules), it is characterized by sub-quadratic space complexity, and it does not incorporate any sparsity pattern priors. Furthermore, it provides strong theoretical guarantees: unbiased estimation of the regular attention matrix and uniform convergence. FAVOR is designed for long input sequences where the number of tokens $L$ satisfies $L \gg d$, for embedding dimensionality $d$. In contrast to previous approaches, instead of simplifying regular attention via various structural priors (which can lead to different, potentially incompatible architectures), we show that it can be effectively approximated as it is, without any "liftings". This leads to our method being flexible: combined with small amounts of fine-tuning, the Performer is backwards-compatible with pretrained regular Transformers and can be also used beyond the Transformer scope as a more scalable replacement for regular attention, which itself has a wide variety of uses in computer vision \cite{attention_cvpr}, reinforcement learning \cite{relational}, and even combinatorial optimization \cite{pointer}. We demonstrate its effectiveness on challenging tasks that include protein sequence modeling and ImageNet64.

We show that regular attention can be considered a special 
case of a much larger class of kernel-driven attention mechanisms, \textit{Generalized Attention} (GA), and that all our results for regular attention can be directly translated also to this extended class. This observation enables us to explore a much larger class of attention models (Sec. \ref{sec:generalized}). Interestingly, 
this is often enabled by the FAVOR mechanism, even if linear scaling is not required (Sec. \ref{sec:experiments}).  We highlight the following contributions: 
%
\begin{itemize}
\item We present \textit{Fast Attention Via Orthogonal Random features} (FAVOR) (Sec. \ref{sec:algorithm}) which can be used as a replacement for regular attention. FAVOR is characterized by $O(Ld\log(d))$ space complexity and $O(Ld^{2}\log(d))$ time complexity, as compared to $O(L^{2})$ space complexity and $O(L^{2}d)$ time complexity for the regular algorithm (Sec. \ref{sec:time_complexity}, Sec. \ref{sec:theory}). 
\item We present a general class of kernel-based attention mechanisms, \textit{Generalized Attention} (GA), which can be handled by FAVOR. Standard attention is a special case. (Sec. \ref{sec:generalized}).
\item We provide strong theoretical guarantees regarding FAVOR: unbiasedness of our estimator of the attention matrix (Sec. \ref{sec:rfm}) and uniform convergence (Sec. \ref{sec:theory})
\item We empirically compare the performance of FAVOR via $\mathrm{Performers}$ at protein sequence modeling tasks, demonstrating in practice all the aforementioned advantages (Sec. \ref{sec:experiments}).
\item We show that our mechanism, implemented in Jax \cite{jax}, is API-compatible with the regular Transformer, whose standard dot-product attention can be replaced by FAVOR with all other components of the architecture intact.
\end{itemize}
All proofs are given in full in the Appendix.

%% file: algorithm.tex
\section{Generalized Attention via FAVOR mechanism}
\label{sec:algorithm}
Below we describe in detail our FAVOR mechanism which is the backbone of our $\mathrm{Performer's}$ architecture. We also present a general class of kernel-based attentions, called \textit{Generalized Attention} (GA) (which includes regular attention as a special case), where FAVOR can be applied.

\subsection{Preliminaries - standard attention mechanism}

Let $L$ be the size of an input sequence of tokens. Then regular dot-product attention \cite{transformer} is a mapping which accepts matrices $\mathbf{Q}, \mathbf{K}, \mathbf{V} \in \mathbb{R}^{L \times d}$ as input where $d$ is the hidden dimension (dimension of the latent representation). Matrices $\mathbf{Q}, \mathbf{K}, \mathbf{V}$ are intermediate representations of the input and their rows can be interpreted as \textit{queries}, \textit{keys} and \textit{values} of the continuous dictionary data structure respectively. \textit{Bidirectional (or non-directional \cite{bert}) dot-product attention} has the following form:
\begin{equation}
    \mathrm{Att}_\leftrightarrow (\mathbf{Q}, \mathbf{K}, \mathbf{V}) = \mathbf{D}^{-1} \mathbf{A} \mathbf{V}, \quad 
    \mathbf{A} = \exp ( \mathbf{Q} \mathbf{K}^\top / \sqrt{d}), \quad \mathbf{D} = \mathrm{diag} ( \mathbf{A} \mathbf{1}_L ) , \label{eq:attnorm}
\end{equation}    
where $\exp (\cdot)$ is applied elementwise, $\mathbf{1}_L$ is the all-ones vector of length $L$, and $\mathrm{diag} (\cdot)$ is a diagonal matrix with the input vector as the diagonal. The runtime complexity of computing (\ref{eq:attnorm}) is $O(L^2 d)$ because the attention matrix $\mathbf{A} \in \mathbb{R}^{L \times L}$ has to be computed and stored explicitly. Hence, in principle, dot-product attention of type (\ref{eq:attnorm}) is incompatible with end-to-end processing of long sequences.

Another important type of attention is \textit{unidirectional dot-product attention} which has the form:
\begin{equation}
    \mathrm{Att}_\to (\mathbf{Q}, \mathbf{K}, \mathbf{V}) = \widetilde{\mathbf{D}}^{-1} \widetilde{\mathbf{A}} \mathbf{V}, \quad 
    \widetilde{\mathbf{A}} = \mathrm{tril} (\mathbf{A}), \quad \widetilde{\mathbf{D}} = \mathrm{diag} ( \widetilde{\mathbf{A}} \mathbf{1}_L ) , \label{eq:uattnorm}
\end{equation}
where $\mathrm{tril}(\cdot)$ returns the lower-triangular part of the argument matrix including diagonal. As discussed in \cite{transformer}, unidirectional attention is used for autoregressive generative modelling with Transformers when the output sequence $o_1, \dots, o_L$ is modelled as:
\begin{equation*}
    p(o_1, \dots, o_L) = p(o_1) p(o_2 | o_1) \dots p(o_L | o_1, \dots, o_{L - 1}) .
\end{equation*}
Therefore, the probability distribution over $o_i$ can only depend on embeddings of tokens $o_1, \dots, o_{i - 1}$. Unidirectional attention is used as self-attention in generative Transformers as well as the decoder part of Seq2Seq Transformers \cite{transformer}, while bidirectional attention is used in encoder self-attention and encoder-decoder attention in Seq2Seq architectures.

A line of work relies on sparse approximation of the matrix $\mathbf{A}$ -- either through restricting the sparsity pattern of $A$ \cite{sparsetr} or learning it using Locality-Sensitive Hashing (LSH) techniques \cite{reformer}. The latter results in $O(L d^2 \log L)$ runtime complexity. We will show that, without any structural assumptions, the matrix $\mathbf{A}$ can be approximated up to any precision in time $O (L d^2\log(d))$.

\subsection{Generalized Attention (GA)}
\label{sec:generalized}
The idea of the attention mechanism is simple. New representations of tokens are obtained from previous ones by taking convex combinations of different value vectors with coefficients of the convex combinations interpreted as renormalized (i.e. all coefficients sum up to one) \textit{similarity measures} between different tokens.
High similarities imply strong attendance to the corresponding tokens. These similarity measures $\mathrm{sim}:\mathbb{R}^{d} \times \mathbb{R}^{d} \rightarrow \mathbb{R}$ are simple ad-hoc ``soft-max style" functions of a dot-product between query $\mathbf{Q}_{i}$ of token $i$ and key $\mathbf{K}_{j}$ of token $j$, namely:
\begin{equation}
\label{standard}
\mathrm{sim}(o_{i},o_{j}) = \mathrm{exp}\left(\frac{\mathbf{Q}_{i}\mathbf{K}_{j}^{\top}}{\sqrt{d}}\right),    
\end{equation}
where: $\mathbf{Q}_{i}^{\top},\mathbf{K}_{j}^{\top} \in \mathbb{R}^{d}$. Note that $\mathrm{sim}$ is not a commutative operation here, and the $\sqrt{d}$-renormalizer is a technical modification to stabilize the range of $\mathrm{sim}$ and avoid very small/large values.

However, what if we use kernels instead of arbitrary similarity measures? Specifically, $\mathbf{Q}_{i}$ and $\mathbf{K}_{j}$ are entangled through a valid kernel function, by defining the attention matrix $\mathbf{A}$ as:
\begin{equation}
\label{general}
\mathbf{A} = \mathbf{A}^{g, h}_{K} = [g(\mathbf{Q}_{i}^{\top})K(\mathbf{Q}_{i}^{\top}, \mathbf{K}_{j}^{\top})h(\mathbf{K}_{j}^{\top})]_{i,j \in \{1,...,L\}},  
\end{equation}
where $K:\mathbb{R}^{d} \times \mathbb{R}^{d} \rightarrow \mathbb{R}$ is an arbitrary kernel function and $g, h:\mathbb{R}^{d} \rightarrow \mathbb{R}$. We call this attention mechanism defined above \textit{Generalized Attention} (GA) parameterized by $K, g, h$.

Next we show that not only can FAVOR approximate regular attention governed by Eq. \ref{standard}, but it can be applied to GAs as long as the corresponding kernels can be effectively estimated via a random feature map mechanism \cite{fourierapprox}, which is the case for most kernels used in practice. We will in fact show that regular attention is a special case of GA for a specific choice of $g, h$, and Gaussian kernel $K$. 

\subsection{Towards FAVOR: approximating attention with random features (RFs)}
\label{sec:rfm}

%

Instead of computing and storing the attention matrix $\mathbf{A} \in \mathbb{R}^{L \times L}$ explicitly, we derive its unbiased stochastic approximation, which benefits from low-rank structure. We take our inspiration from a randomized scheme to train kernel Support Vector Machines with large training data \cite{fourierapprox}. 

Let $\mathbf{Q}_i$ and $\mathbf{K}_i$ denote the $i$-th rows of matrices $\mathbf{Q}$ and $\mathbf{K}$ respectively. For regular attention, the $i,j$-th element of $\mathbf{A}$ can be expressed as:
\begin{equation*}
    \mathbf{A}_{i,j} = \exp (\mathbf{Q}_i \mathbf{K}_j^{\top} / \sqrt{d}) = \exp (\| \mathbf{Q}_i \|^2_2 / 2\sqrt{d}) \cdot \exp (- \| \mathbf{Q}_i - \mathbf{K}_j \|^2_2 / 2\sqrt{d}) \cdot \exp ( \| \mathbf{K}_j \|_2^2 / 2\sqrt{d}).
\end{equation*}
In other words, for $r=2\sqrt{d}$, the attention matrix $\mathbf{A}$ can be decomposed as:
\begin{gather}
    \mathbf{A} = \mathbf{D}_\mathbf{Q} \mathbf{B} \mathbf{D}_\mathbf{K}, \quad \mathbf{B} \in \mathbb{R}^{L \times L}, \forall i,j: \mathbf{B}_{i,j} = \exp (- \| \mathbf{Q}_i - \mathbf{K}_j \|^2_2 / r), \label{eq:b} \\
    \mathbf{D}_\mathbf{T} = \mathrm{diag} \biggl( \exp(\| \mathbf{T}_1 \|_2^2 / r), \dots, \exp(\| \mathbf{T}_L \|_2^2 / r) \biggr),  \label{eq:dqdk}
\end{gather}
for $\mathbf{T}=\mathbf{Q},\mathbf{K}$. Both $\mathbf{D}_\mathbf{Q}$ and $\mathbf{D}_\mathbf{K}$ can be computed in $O(Ld)$ time. 
Note that the $i,j$-th element of matrix $\mathbf{B}$ is the value of the Gaussian kernel with $\sigma = d^{\frac{1}{4}}$: 
\begin{equation}
\mathbf{B}_{i,j} = K^{\sigma}_{\mathrm{gauss}}(\mathbf{Q}_{i}^{\top}, \mathbf{K}_{j}^{\top}) 
\overset{\mathrm{def}}{=} \mathrm{exp}(-\frac{\|\mathbf{Q}_{i}-\mathbf{K}_{j}\|^{2}}{2\sigma^{2}}).
\end{equation}
For GA, our analysis is similar. This time $\mathbf{D}_{\mathbf{Q}}, \mathbf{D}_{\mathbf{K}}$ have nonzero entries of the form $g(\mathbf{Q}_{i}^{\top})$ and $h(\mathbf{K}_{i}^{\top})$ (for regular attention we have: $g(\mathbf{x})=h(\mathbf{x})=\exp(\frac{\|\mathbf{x}\|_{2}^{2}}{r})$) respectively and furthermore the Gaussian kernel is replaced by a general kernel $K$, namely: 
$\mathbf{B}_{i,j}=K(\mathbf{Q}_{i}^{\top},\mathbf{K}_{j}^{\top})$, as in Equation \ref{general}. 

In the reminder of this section we will derive an unbiased stochastic approximation of matrix $\mathbf{B}$ based on low-rank decomposition of $\mathbf{B}$ with the use of random feature maps \cite{fourierapprox}. 

For a given kernel $K:\mathbb{R}^{d} \times \mathbb{R}^{d} \rightarrow \mathbb{R}$, the random feature [RF] map $\phi_{K}: \mathbb{R}^{d} \rightarrow \mathbb{R}^{M}$ corresponding to $K$ is a probabilistic embedding satisfying
\begin{equation}
\label{rfm}
K(\mathbf{x}, \mathbf{y}) = \mathbb{E}[\phi^{\top}(\mathbf{x})\phi(\mathbf{y})] ,
\end{equation}
where the expectation is with respect to the randomness of $\phi$, and $M$ denotes the number of random features (if $\mathbb{E}[\phi^{\top}(\mathbf{x})\phi(\mathbf{y})]$ only approximates $K(\mathbf{x}, \mathbf{y})$ then we refer to the mechanism as an \textit{approximate random feature map}).
This mechanism covers also as a very special case an instance, where $\phi$ is deterministic (and then expectation is not needed).
Efficient-to-compute random feature maps exist for virtually all classes of kernels used in machine learning, e.g. shift-invariant kernels \cite{fourierapprox}, the pointwise nonlinear Gaussian kernel related to neural networks \cite{gulrajani}, and more, though the techniques used to derive these random mappings vary from class to class \cite{unreas}. 
Even more interestingly, for most of these kernels, corresponding random feature maps have a similar structure, namely: 
\begin{equation}
\label{interesting}
\phi(\mathbf{x})\overset{\mathrm{def}}{=}\frac{c}{\sqrt{M}}
(f(\omega_{1}^{\top}\mathbf{x}+b_{1}),...,f(\omega_{M}^{\top}\mathbf{x}+b_{M}))^{\top}=\frac{c}{\sqrt{M}}f(\mathbf{Wx}+\mathbf{b})^{\top},
\end{equation} 
for some $f:\mathbb{R} \rightarrow \mathbb{R}$, $\omega_{1},...,\omega_{M} \overset{\mathrm{iid}}{\sim} \Omega$, $b_{1},...,b_{M} \overset{\mathrm{iid}}{\sim} \mathcal{B}$, distributions: $\Omega \in \mathcal{P}(\mathbb{R}^{d})$, $\mathcal{B} \in \mathcal{P}(\mathbb{R})$ and constant $c>0$.
Here $\mathbf{W} \in \mathbb{R}^{M \times d}$ has rows $\mathbf{W}_{i} = \omega_{i}^{\top}$ and $\mathbf{b} \overset{\mathrm{def}}{=} (b_{1},...,b_{M})^{\top}$.

In particular, for the Gaussian kernel, we have $c=\sqrt{2}$ and:
\begin{equation}
\phi(\mathbf{x}) \overset{\mathrm{def}}{=}
\sqrt{\frac{2}{M}}(\cos(\omega^{\top}_{1}\mathbf{x}+b_{1}),...,\cos(\omega^{\top}_{M}\mathbf{x}+b_{M}))^{\top},
\end{equation}
where $\omega_{1},...,\omega_{M} \overset{\mathrm{iid}}{\sim} \mathcal{N}(0,\sigma^{2}\mathbf{I}_{d})$ and $b_{1},...,b_{M} \overset{\mathrm{iid}}{\sim} \mathrm{Unif}(0, 2\pi)$. This particular form of $\phi$ is a consequence of the celebrated Bochner's Theorem \cite{fourierapprox}.
We now define $\widehat{\mathbf{Q}}$ and $\widehat{\mathbf{K}} \in \mathbb{R}^{L \times M}$ as: 
\begin{equation}
\label{qkprime}
\widehat{\mathbf{Q}} = \frac{c}{\sqrt{M}}f(\mathbf{W}\mathbf{Q}^{\top}+\mathbf{b})^{\top}, \quad   
\widehat{\mathbf{K}} = \frac{c}{\sqrt{M}}f(\mathbf{W}\mathbf{K}^{\top}+\mathbf{b})^{\top}.
\end{equation}
Note that we have: $\widehat{\mathbf{Q}}_{i} = \phi(\mathbf{Q}_{i}^{\top})^{\top}$ and $\widehat{\mathbf{K}}_{i} = \phi(\mathbf{K}_{i}^{\top})^{\top},$ 
where $\widehat{\mathbf{Q}}_{i}$ and $\widehat{\mathbf{K}}_{i}$ stand for the ith row of $\widehat{\mathbf{Q}}$ and $\widehat{\mathbf{K}}$ respectively. Then according to Equation \ref{rfm}, we have: $\mathbf{B} = \mathbb{E}[\widehat{\mathbf{Q}}\widehat{\mathbf{K}}^{\top}]$.
Thus with $\mathbf{Q}^{\prime}$, $\mathbf{K}^{\prime}$ given as:
$\mathbf{Q}^{\prime} = \mathbf{D}_{\mathbf{Q}}\widehat{\mathbf{Q}}$,
$\mathbf{K}^{\prime} = \mathbf{D}_{\mathbf{K}}\widehat{\mathbf{K}}$, we obtain:
\begin{equation}
\label{unbiased_attention_estimation}  
\mathbf{A} = \mathbb{E}[\mathbf{Q}^{\prime}(\mathbf{K}^{\prime})^{\top}].
\end{equation}
We conclude that the attention matrix $\mathbf{A}$ can be approximated without bias as: $\widehat{\mathbf{A}}=\mathbf{Q}^{\prime}(\mathbf{K}^{\prime})^{\top}$.
We will leverage this unbiased approximate low-rank (if $M \ll L$) decomposition of $\mathbf{A}$ in our algorithm, even though we will not explicitly compute $\widehat{\mathbf{A}}$.

Note that one can also define a valid kernel as: 
$K(\mathbf{x},\mathbf{y}) = \mathbb{E}[\phi(\mathbf{x})^{\top}\phi(\mathbf{y})]$
for $\phi$ as in Eq. \ref{interesting} and an arbitrary $f:\mathbb{R} \rightarrow \mathbb{R}$. Such kernels cover in particular the family of \textit{Pointwise Nonlinear Gaussian Kernels} \cite{unreas} (intrinsically related to nonlinear neural networks) such as arc-cosine kernels (e.g. angular kernels). Most of these kernels do not have closed-forms so computing exact GAs for them would not be possible, but of course computation is feasible with the presented mechanism.

\subsection{Towards FAVOR: refinements via orthogonal random features}
\label{sec:orfms}
For isotropic $\Omega$ (true for most practical applications, including regular attention), instead of sampling $\omega_{i}$ independently, we can use \emph{orthogonal random features} (ORF) \cite{ort, unreas, geom}: these maintain (exactly or approximately) the marginal distributions of samples $\omega_{i}$ while enforcing that different samples are orthogonal. If we need $M>d$, ORFs still can be used locally within each $d \times d$ block of $\mathbf{W}$ \cite{ort}.

ORFs were introduced to reduce the variance of Monte Carlo estimators \cite{ort, unreas, geom, kama, hron,psrnn, uni} and we show in Secs. \ref{sec:theory} and \ref{sec:experiments} that they do indeed lead to more accurate approximations and substantially better downstream results. Below we breifly review the most efficient ORF mechanisms (based on their strengths and costs) that we will use in Sec. \ref{sec:time_complexity} in the analysis of FAVOR.

\textbf{(1) Regular ORFs [R-ORFs]:} Applies Gaussian orthogonal matrices \cite{ort}. Encodes matrix $\mathbf{W}$ in $O(Md)$ space. Provides algorithm for computing $\mathbf{Wx}$ in $O(Md)$ time for any $\mathbf{x} \in \mathbb{R}^{d}$. Gives unbiased estimation. Requires one-time $O(Md^{2})$ preprocessing (Gram-Schmidt orthogonalization).

\textbf{(2) Hadamard/Givens ORFs [H/G-ORFs]:} Applies random Hadamard \cite{unreas}/Givens matrices \cite{uni}. Encodes matrix $\mathbf{W}$ in $O(M)$/$O(M\log(d))$ space. Provides algorithm for computing $\mathbf{Wx}$ in $O(M\log(d))$ time for any $\mathbf{x} \in \mathbb{R}^{d}$. Gives small bias (going to $0$ with $d \rightarrow \infty$).

\subsection{FAVOR: Fast Attention via Orthogonal Random features} 

We are ready to present the full FAVOR algorithm. In the bidirectional case, our approximate attention computed by FAVOR is given as: 
\begin{equation}
\widehat{\mathrm{Att}}_\leftrightarrow (\mathbf{Q}, \mathbf{K}, \mathbf{V}) = \widehat{\mathbf{D}}^{-1} \widehat{\mathbf{A}} \mathbf{V}=
\widehat{\mathbf{D}}^{-1}(\mathbf{Q}^{\prime}((\mathbf{K}^{\prime})^{\top}\mathbf{V})),
\end{equation}

where $\widehat{\mathbf{D}} = \mathrm{diag}(\mathbf{Q}^{\prime}((\mathbf{K}^{\prime})^{\top}\mathbf{1}_{L}))$. The placement of brackets determines the order in which computations are conducted. Note that we never explicitly compute $\widehat{\mathbf{A}}$ and consequently, avoid $\Theta(L^2)$ time complexity and storing the $L \times L$ approximate attention matrix (see: Sec. \ref{sec:time_complexity} for rigorous analysis).
\begin{algorithm}[t]
\SetAlgoLined
\SetKwInOut{Input}{Input}
\Input{ $\mathbf{Q}, \mathbf{K}, \mathbf{V} \in \mathbb{R}^{L \times d}$, $\mathrm{isBidirectional}$ - binary flag.}
\KwResult{$\widehat{\mathrm{Att}}_\leftrightarrow (\mathbf{Q}, \mathbf{K}, \mathbf{V}) \in \mathbb{R}^{L \times L}$ if $\mathrm{isBidirectional}$, $\widehat{\mathrm{Att}}_\to (\mathbf{Q}, \mathbf{K}, \mathbf{V}) \in \mathbb{R}^{L \times L}$ otherwise.}
Compute $\mathbf{D}_\mathbf{Q}, \mathbf{D}_\mathbf{K}$ as explained in Sec. \ref{sec:rfm}\;
Compute $\widehat{\mathbf{Q}}, \widehat{\mathbf{K}}$ according to (\ref{qkprime}) and take $\mathbf{Q}^{\prime} := \mathbf{D}_\mathbf{Q} \widehat{\mathbf{Q}}, \quad \mathbf{K}^{\prime} := \mathbf{D}_\mathbf{K} \widehat{\mathbf{K}}, \quad \mathbf{C} := \begin{bmatrix} \mathbf{V} & \mathbf{1}_L \end{bmatrix}$\;
\eIf{$\mathrm{isBidirectional}$}{
    $\mathrm{Buf}_1 := (\mathbf{K}^{\prime})^\top \mathbf{C} \in \mathbb{R}^{M \times (d + 1)}, \quad \mathrm{Buf}_2 := \mathbf{Q}^{\prime} \mathrm{Buf}_1 \in \mathbb{R}^{L \times (d + 1)}$\;
}{
    Compute $\mathbf{G}$ and its prefix-sum tensor $\mathbf{G}^\mathrm{PS}$ according to (\ref{eq:cumsum})\;
    $\mathrm{Buf}_2 := \begin{bmatrix} \mathbf{G}^\mathrm{PS}_{1,:,:} \mathbf{Q}^{\prime}_1 & \dots & \mathbf{G}^\mathrm{PS}_{L,:,:} \mathbf{Q}^{\prime}_L \end{bmatrix}^\top \in \mathbb{R}^{L \times (d + 1)}$\;
}
$\begin{bmatrix} \mathrm{Buf}_3 & \mathrm{buf}_4 \end{bmatrix} := \mathrm{Buf}_2, \quad \mathrm{Buf}_3 \in \mathbb{R}^{L \times d}, \quad \mathrm{buf}_4 \in \mathbb{R}^L$\;
\Return $\mathrm{diag} (\mathrm{buf}_4)^{-1} \mathrm{Buf}_3$\;
\caption{FAVOR (bidirectional or unidirectional).}
\label{alg:1}
\end{algorithm}

\subsubsection{Prefix-sums for unidirectional FAVOR}

For the unidirectional case, our analysis is similar but this time our goal is to compute $\mathrm{tril}(\mathbf{Q}^{\prime} (\mathbf{K}^{\prime})^\top) \mathbf{C}$ without constructing and storing the $L \times L$-sized matrix $\mathrm{tril}(\mathbf{Q}^{\prime} (\mathbf{K}^{\prime})^\top)$ explicitly, where
$C~=~\begin{bmatrix} V & \mathbf{1}_L \end{bmatrix} \in \mathbb{R}^{L \times (d + 1)}$. In order to do so, observe that $\forall 1 \leq i \leq L$:
\begin{equation} \label{eq:cumsum}
    [\mathrm{tril}(\mathbf{Q}^{\prime} (\mathbf{K}^{\prime})^\top) \mathbf{C}]_i = \mathbf{G}^\mathrm{PS}_{i,:,:} \times \mathbf{Q}^{\prime}_i, \quad \mathbf{G}^\mathrm{PS}_{i,:,:} = \sum_{j = 1}^i \mathbf{G}_{j,:,:}, \quad \mathbf{G}_{j,:,:} = \mathbf{K}^{\prime}_j \mathbf{C}_j^\top \in \mathbb{R}^{M \times (d + 1)}
\end{equation}
where $\mathbf{G}, \mathbf{G}^\mathrm{PS} \in \mathbb{R}^{L \times M \times (d + 1)}$ are 3d-tensors. Each slice $\mathbf{G}^\mathrm{PS}_{:,l,p}$ is therefore a result of a prefix-sum (or cumulative-sum) operation applied to $\mathbf{G}_{:,l,p}$: $\mathbf{G}^\mathrm{PS}_{i,l,p} = \sum_{j = 1}^i \mathbf{G}_{i,l,p}$. An efficient algorithm to compute the prefix-sum of $L$ elements takes $O(L)$ total steps and $O(\log L)$ time when computed in parallel \cite{cumsum, cormen}. 
See Algorithm \ref{alg:1} for the whole approach.

\subsection{Time and space complexity analysis}
\label{sec:time_complexity}
We see that a variant of bidirectional FAVOR using regular RFs (based on iid samples) or R-ORFs has $O(Md+Ld+ML)$ space complexity as opposed to $\Theta(L^{2} + Ld)$ space complexity of the baseline. Unidirectional FAVOR using fast prefix-sum precomputation in parallel \cite{cumsum, cormen} has $O(M L d)$ space complexity to store $\mathbf{G}^\textrm{PS}$ which can be reduced to $O(Md+Ld+ML)$ by running a simple (though non-parallel in $L$) aggregation of $\mathbf{G}^\textrm{PS}_{i,:,:}$ without storing the whole tensor $\mathbf{G}^{PS}$ in memory. From Sec. \ref{sec:orfms}, we know that if instead we use G-ORFs, then space complexity is reduced to $O(M\log(d) + Ld + ML)$ and if the H-ORFs mechanism is used, then space is further reduced to $O(M + Ld + ML)=O(Ld+ML)$. Thus for $M,d \ll L$ all our variants provide substantial space complexity improvements since they do not need to store the attention matrix explicitly.

The time complexity of Algorithm \ref{alg:1} is $O(L M d)$ (note that constructing $\widehat{\mathbf{Q}}$ and $\widehat{\mathbf{K}}$ can be done in time $O(LMd)$ via Eq. \ref{qkprime} if samples from $\Omega$ and $\mathcal{B}$ can be obtained in time $O(d)$ and $O(1)$ respectively (which is the case for all practical applications). Note that the time complexity of our method is much lower than $O(L^2 d)$ of the baseline for $L \gg M$.

As explained in Sec. \ref{sec:orfms}, the R-ORF mechanism incurs an extra one-time $O(Md^{2})$ cost (negligible compared to the $O(LMd)$ term for $L \gg d$). H-ORFs or G-ORFs do not have this cost, and when FAVOR uses them, computing $\mathbf{Q}^{\prime}$ and $\mathbf{K}^{\prime}$ can be conducted in time $O(L\log(M)d)$ as opposed to $O(LMd)$ (see: Sec. \ref{sec:orfms}). Thus even though H/G-ORFs do not change the asymptotic time complexity, they improve the constant factor from the leading term. This plays an important role for training very large models. 

The number of random features $M$ allows a trade-off between computational complexity and the level of approximation: bigger $M$ results in higher computation costs, but also in a lower variance of the estimate of $\mathbf{A}$. In the next section we will show that in practice we can take $M=\Theta(d\log(d))$.

Observe that the algorithm obtained is highly-parallelizable, and benefits from fast matrix multiplication and broadcasted operations on GPUs or TPUs.

%% file: theory.tex
\section{Theoretical convergence analysis}
\label{sec:theory}
In contrast to other methods approximating the attention matrix $\mathbf{A}$, our algorithm provides provable strong uniform convergence theoretical guarantees for compact domains. We show that $M_{\mathrm{opt}}$, the optimal number of random features, does not depend on $L$ but only on $d$. In fact, we prove that if we take $M_{\mathrm{opt}} = \Theta(d\log(d))$, then with $O(L d^2\log(d))$-time, we can approximate $\mathbf{A}$ up to any precision, regardless of the number of tokens $L$. In order to provide those guarantees for FAVOR, we leverage recent research on the theory of negative dependence for ORFs \cite{Lin2020DemystifyingOM}.
The following is true:

\begin{theorem}[Uniform convergence of FAVOR]
\label{uniform_convergence}
Take the generalized attention mechanism defined by $g,h:\mathbb{R}^{d} \rightarrow \mathbb{R}$ (see: Sec. \ref{sec:generalized}) and a radial basis function (RBF) kernel \cite{geom} $K$ with corresponding spectral distribution $\Omega$ (e.g. Gaussian kernel for which $\Omega=\mathcal{N}(0,\mathbf{I}_{d})$). Assume that the rows of matrices $\mathbf{Q}$ and $\mathbf{K}$ are taken from a ball $B(R)$ of radius $R$, centered at $0$ (i.e. norms of queries and keys are upper-bounded by $R$). 
Define $l=Rd^{-\frac{1}{4}}$ and take $g^{*} = \max_{\mathbf{x} \in B(l)}|g(\mathbf{x})|$, $h^{*} = \max_{\mathbf{x} \in B(l)}|h(\mathbf{x})|$.
Then for any $\epsilon>0$, $\delta=\frac{\epsilon}{g^{*}h^{*}}$
and the number of random features $M = \Omega(\frac{d}{\delta^{2}}\log(\frac{4\sigma R}{\delta d^{\frac{1}{4}}}))$ for $\sigma=\mathbb{E}_{\omega \sim \Omega}[\omega^{\top}\omega]$ the following holds:
$
\|\widehat{\mathbf{A}}-\mathbf{A}\|_{1} \leq \epsilon     
$
with any constant probability,
where $\widehat{\mathbf{A}}$ approximates generalized attention matrix via FAVOR with R-ORFs. 
\end{theorem}

The result holds in particular for regular attention using Gaussian kernels (see: Sec. \ref{sec:generalized}) for which $M_{\mathrm{opt}}=\Omega(\frac{d}{\delta^{2}}\log(\frac{4d^{\frac{3}{4}} R}{\delta}))$ since $\sigma = d$.

%% file: experiments.tex
\section{Experiments}
\label{sec:experiments}
We implement our setup on top of pre-existing Transformer training code in Jax \cite{jax} optimized with just-in-time (\texttt{jax.jit}) compilation, and complement our theory with empirical evidence to demonstrate FAVOR's practicality in the protein setting. Unless explicitly stated, a Performer replaces only the attention component with FAVOR, while all other components are exactly the same as for the regular Transformer. Furthermore, since we use the cross-entropy loss in our generative training experiments, we use the standard accuracy metric as defined from supervised learning. 

\subsection{Computation costs}
\begin{figure}[h]
  \centering
  \includegraphics[width=1.0\textwidth]{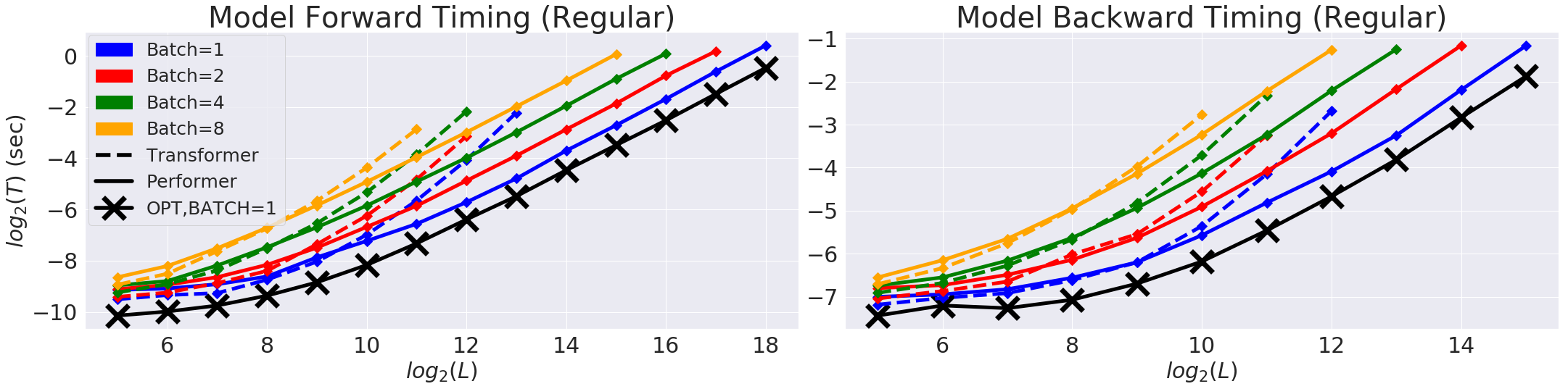}
  \caption{\small{Comparison of Transformer and Performer in terms of forward and backward pass speed and maximum $L$ allowed. "X" (OPT) denotes the maximum possible speedup achievable, when attention simply returns the $\mathbf{V}$-vector. Plots shown up to when a model produces an out of memory error on a V100 GPU with 16GB. Best in color.}}
  \label{fig:backward_pass}
\end{figure}

Between the Transformer and the Performer, we compared speed-wise the backward pass, as it is one of the main computational bottlenecks during training, when using the regular default size $(n_{heads}, n_{layers}, d_{ff}, d) = (8,6,2048,512)$, where $d_{ff}$ denotes the width of the MLP layers. We observed (Fig. \ref{fig:backward_pass}) that in terms of $L$, the Performer reaches nearly linear time complexity as opposed to the Transformer's quadratic time complexity. The Performer's memory consumption is also sub-quadratic (as it does not store the explicit $O(L^{2})$-sized attention matrix), allowing higher batch sizes and longer sequence lengths. In fact, the Performer achieves nearly optimal speedup and memory efficiency possible, depicted by the "X"-line when attention is replaced a "identity function" by simply returning the $\mathbf{V}$-vector. The combination of both memory and backward pass efficiencies for large $L$ has profound implications for training speed, as it allows respectively, large batch training and lower wall clock time per gradient step, contributing to total train time reduction. Extensive additional results are demonstrated in Appendix \ref{appendix:computation_costs_bidirectional} by varying layers, raw attention, and architecture sizes.

\subsection{Approximation error and compatibility with regular Transformer}
\label{subsec:approx_error_compatibility}
We further examined the approximation error of the attention matrix implicitly defined in FAVOR in Fig. \ref{fig:approx} (and in Fig. \ref{fig:appendix_approx} in Appendix \ref{appendix:extended_approx}), which thus directly affects the accuracy of FAVOR's output. 
We demonstrate that orthogonal features generally produce lower error than unstructured features. 
\vspace{0.5mm}
\begin{figure}[h]
  \includegraphics[width=0.498\textwidth]{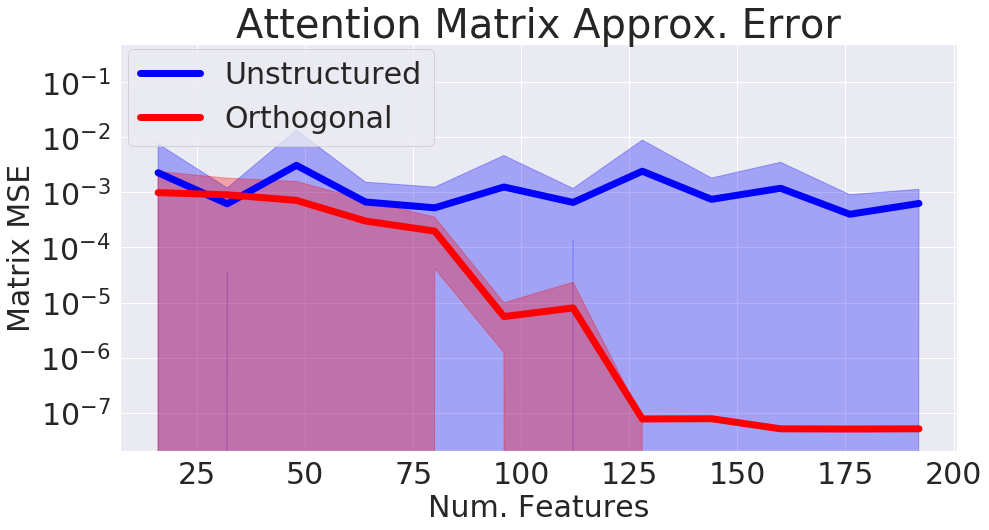}
  \includegraphics[width=0.498\textwidth]{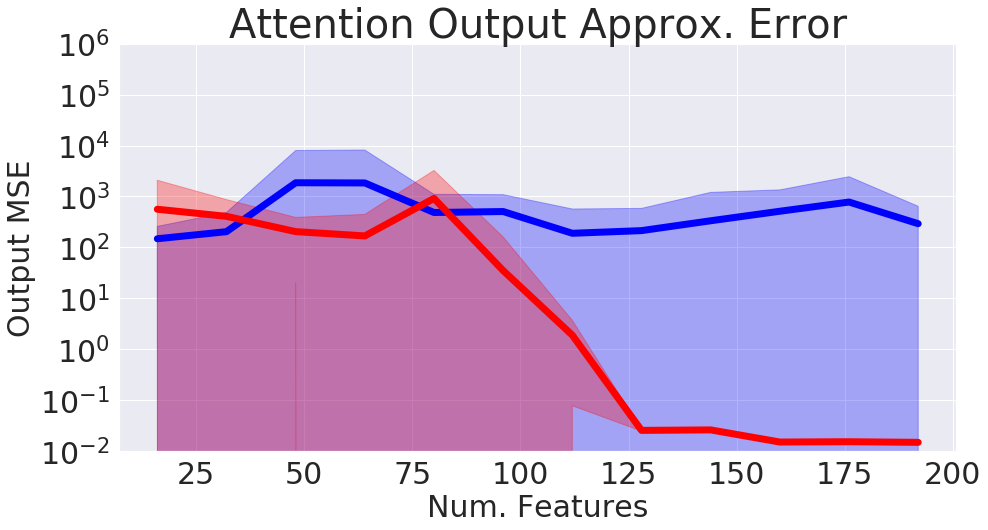}
  \caption{\small{Approximation errors for both the attention matrix and output of the mechanism itself. We took $L = 4096, d = 16$, and varied the number of random features $M$. Standard deviations shown across 10 samples.}}
  \label{fig:approx}
\end{figure}

\begin{figure}[h]
  \centering
  \includegraphics[width=0.65\textwidth]{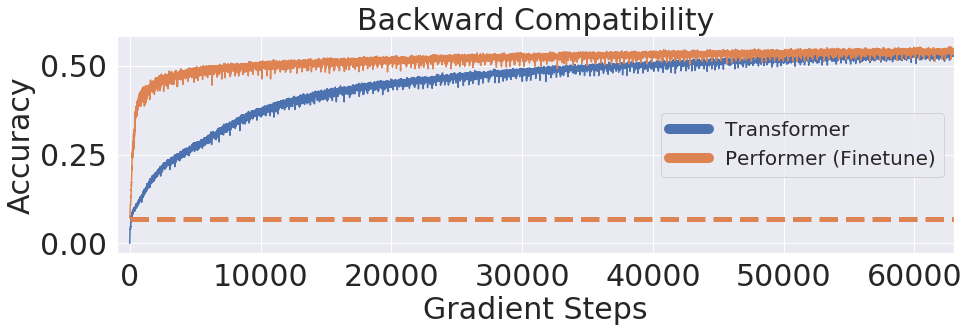}
  \caption{\small We transferred the original pretrained Transformer's weights into the Performer, which produces an initial non-zero 0.07 accuracy (dotted orange line). Once fine-tuned however, the Performer quickly recovers accuracy in a small fraction of the original number of gradient steps.}
  \label{fig:backward_compatibility}
\end{figure}

Notice that the accuracy can be further boosted by applying a resampling strategy that reconstructs samples periodically. We set this option as a hyperparameter of our overall algorithm.

The approximation error can propagate when applying the other components (MLPs, multiple heads, multiple layers, etc.) of a Transformer, which we demonstrate in Fig. \ref{fig:appendix_approx} (Appendix). This implies we cannot immediately directly transfer the weights from a pretrained Transformer onto the Performer. However, this can be resolved by finetuning the Performer on the trained task. We demonstrate this technique for a pretrained BERT model \cite{bert} on the LM1B dataset \cite{lm1b} in Fig. \ref{fig:backward_compatibility}.

\subsection{Multiple layer training}
\begin{figure}[h]
  \centering
  \includegraphics[width=1.0\textwidth]{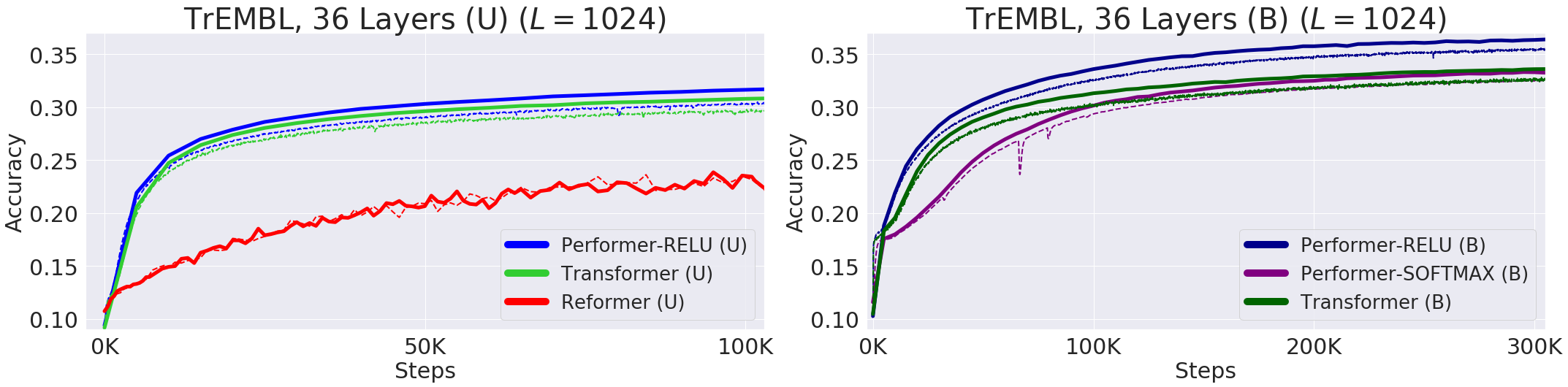}
  \caption{\small Train = Dashed, Validation = Solid, Unidirectional = (U), Bidirectional = (B). For TrEMBL, we used the exact same model parameters $(n_{heads}, n_{layers}, d_{ff}, d)  = (8, 36, 1024, 512)$ from \protect\cite{progen} for all runs. For fairness, all TrEMBL experiments used 16x16 TPU-v2's. Batch sizes were maximized for each separate run given the corresponding compute constraints. Hyperparameters can be found in Appendix \ref{appendix:hyperparameters}.}
  \label{fig:big_benchmarking}
\end{figure}

We further benchmark the Performer on both unidirectional (U) and bidirectional (B) cases by training a 36-layer model using protein sequences from the Jan. 2019 release of TrEMBL \cite{uniprot2019uniprot}, similar to \cite{progen}. As a baseline for sparse attention, we also used the Reformer \cite{reformer}. In Fig. \ref{fig:big_benchmarking}, the Reformer \textit{significantly drops in accuracy} on the protein dataset. This suggests that sparse attention may be insufficient for protein tasks, which require modelling of global interactions. Furthermore, the usefulness of generalized attention is evidenced by Performer-RELU (taking $f=\mathrm{RELU}$ in Equation \ref{rfm}) achieving the highest accuracy in both (U) and (B) cases. Our proposed softmax approximation is also shown to be tight, achieving the same accuracies as the exact-softmax Transformer. Extended results including dataset statistics, out of distribution evaluations, and visualizations, can be found in Appendix \ref{sec:protein_extended}.

\vspace{1mm}
\subsection{Large length training}
\begin{figure}[h]
  \centering
  \includegraphics[width=1.0\textwidth]{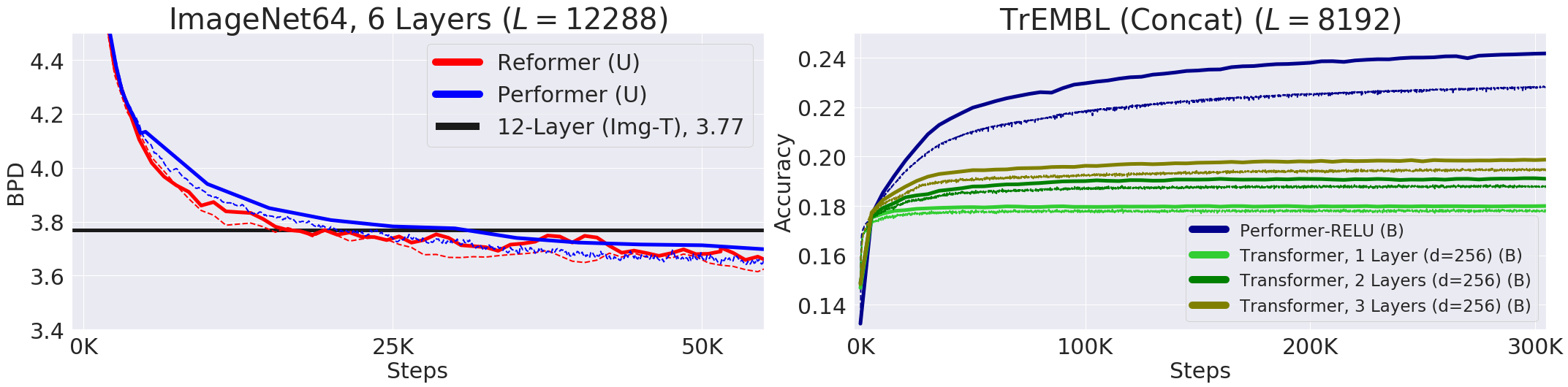}
  \caption{\small 
  For ImageNet64, we took 6-layer variants of both the Performer and the Reformer, evaluated both at 50K steps, and also plotted the 3.77 BPD baseline (Img-T) from \protect\cite{parmar}, which uses 12-layers but with cropped lengths on a regular Transformer. For concatenated TrEMBL, we varied $n_{layers} \in \{1,2,3\}$ for the smaller Transformer. Hyperparameters found in Appendix \ref{appendix:hyperparameters}.}
  \label{fig:im64_protein_8192}
\end{figure}

On the standard (U) ImageNet64 benchmark $(L = 12288)$ from \cite{parmar}, we see that the Performer matches the Reformer, when both use 8x8 TPU-v2's. Depending on hardware (TPU or GPU), we also found that the Performer can be 2x faster than the Reformer via Jax optimizations for the (U) setting. For a proof of principle study, we also create an initial protein benchmark for predicting interactions among groups of proteins by concatenating protein sequences to length $L = 8192$ from TrEMBL, long enough to model protein interaction networks without the large sequence alignments required by existing methods \cite{cong2019protein}. In this setting, a baseline Transformer overloads memory even at a batch size of $1$ per chip, by a wide margin. Thus as a baseline we were forced to use a significantly smaller variant, reducing to $(n_{heads}, n_{layers}, d_{ff}, d) = (8, \{1,2,3\}, 256, 256)$. Meanwhile, the Performer trains efficiently at a batch size of 8 per chip using the standard $(8, 6, 2048, 512)$ architecture. We see in Fig. \ref{fig:im64_protein_8192} that the smaller Transformer ($n_{layer} = 3$) is quickly bounded at $\approx 19 \%$, while the Performer is able to train continuously to $\approx 24\%$.

%% file: conclusion.tex
\section{Conclusion}
\label{sec:conclusion}
We presented $\mathrm{Performer}$, a new type of Transformer, relying on our Fast Attention Via Orthogonal Random features (FAVOR) mechanism to significantly improve space and time complexity of regular Transformers. Our mechanism is to our knowledge the first unbiased estimation of the original algorithm with linear space and time complexity with respect to $L$. Furthermore, FAVOR could be applied to other tasks of approximate attention, including hierarchical attention networks \cite{hans}, graph attention networks \cite{gran}, image processing \cite{attention_cvpr}, and reinforcement learning/robotics \cite{tang}.

%% file: broader_impact.tex
\section{Broader impact}
\label{sec:broader_impact}
We believe that the presented algorithm can be impactful in various ways:

\textbf{Biology and Medicine:} Our method has the potential to directly impact research on biological sequence analysis by enabling the Transformer to be applied to much longer sequences without constraints on the structure of the attention matrix. The initial application that we consider is the prediction of interactions between proteins on the proteome scale. Recently published approaches require large evolutionary sequence alignments, a bottleneck for applications to mammalian genomes \cite{cong2019protein}. The potentially broad translational impact of applying these approaches to biological sequences was one of the main motivations of this work. We believe that modern bioinformatics can immensely benefit from new machine learning techniques with Transformers being among the most promising. Scaling up these methods to train faster more accurate language models opens the door to the ability to design sets of molecules with pre-specified interaction properties. These approaches could be used to augment existing physics-based design strategies that are of critical importance for example in the development of new nanoparticle vaccines \cite{marcandalli2019induction}.  

\textbf{Environment:} As we have shown, Performers with FAVOR are characterized by much lower compute costs and substantially lower space complexity which can be directly translated to $\mathrm{CO}_{2}$ emission reduction \cite{co2} and lower energy consumption \cite{energy}, as regular Transformers require very large computational resources.

\textbf{Research on Transformers:} We believe that our results can shape research on efficient Transformers architectures, guiding the field towards methods with strong mathematical foundations. Our research may also hopefully extend Transformers also beyond their standard scope (e.g. by considering the Generalized Attention mechanism and connections with kernels). Exploring scalable Transformer architectures that can handle $L$ of the order of magnitude few thousands and more, preserving accuracy of the baseline at the same time, is a gateway to new breakthroughs in bio-informatics, e.g. language modeling for proteins, as we explained in the paper. Our presented method can be potentially a first step.

\textbf{Backward Compatibility:} Our Performer can be used on the top of a regular pre-trained Transformer as opposed to other Transformer variants. Even if up-training is not required, FAVOR can be still used for fast inference with no loss of accuracy. We think about this backward compatibility as a very important additional feature of the presented techniques that might be particularly attractive for practitioners.

\textbf{Attention Beyond Transformers:} Finally, FAVOR can be applied to approximate exact attention also outside the scope of Transformers. This opens a large volume of new potential applications including: hierarchical attention networks (HANS) \cite{hans}, graph attention networks \cite{gran}, image processing \cite{attention_cvpr}, and reinforcement learning/robotics \cite{tang}.

%% file: acknowledgements.tex
\section{Acknowledgements}
We thank Afroz Mohiuddin, Wojciech Gajewski, Nikita Kitaev, and Lukasz Kaiser for multiple discussions on the Transformer. We further thank Joshua Meier, Aurko Roy, and John Platt for many fruitful discussions on biological data and useful comments on this draft.  

Lucy Colwell acknowledges support from the Simons Foundation. Adrian Weller acknowledges support from the David MacKay Newton research fellowship at Darwin College, The Alan Turing Institute under EPSRC grant EP/N510129/1 and U/B/000074, and the Leverhulme Trust via CFI.

%% file: appendix.tex
\section*{APPENDIX: Masked Language Modeling for Proteins via Linearly Scalable Long-Context Transformers}
\setcounter{section}{0}
\renewcommand{\thesection}{\Alph{section}}
\label{sec:appendix}

\section{Theoretical results}
\label{appendix:theoretical_results}

We provide here the proof of Theorem \ref{uniform_convergence} from the main body.
\begin{proof}
We consider first the case of the default FAVOR setting with R-ORF mechanism turned on.
We rely on Theorem 3 from \cite{Lin2020DemystifyingOM}.
Note that we can apply it in our case, since for RBF kernels the corresponding function $f$ is $\cos$ (thus in particular it is bounded). Also, it is not hard to observe (see for instance analysis in  Claim 1 from \cite{fourierapprox}) that $L_{f}=1$. 
Using Theorem 3 from \cite{Lin2020DemystifyingOM}, we conclude that:
\begin{equation}
\|\widehat{\mathbf{B}}-\mathbf{B}\|_{1} \leq \delta    
\end{equation}
with any constant probability as long as
$M = \Omega(\frac{d}{\delta^{2}})\log(\frac{\sigma \cdot  \mathrm{diam}(\mathcal{M})}{\delta})$,
where $\sigma=\mathbb{E}[\omega^{\top}\omega]$ and $\mathcal{M}$ is the diameter of the smallest ball $\mathcal{M}$ containing all vectors of the form $\mathbf{z} = \frac{\mathbf{Q}_{i}}{d^{\frac{1}{4}}}-\frac{\mathbf{K}_{j}}{d^{\frac{1}{4}}}$. 
Since $\|\mathbf{Q}_{i}\|_{2}, \|\mathbf{K}_{j}\|_{2} \leq R$, we conclude that $\|\mathbf{z}\|_{2} \leq \frac{2R}{d^{\frac{1}{4}}}$ and thus one can take $\mathrm{diam}(\mathcal{M})=\frac{4R}{d^{\frac{1}{4}}}$.
We have:
\begin{equation}
\|\widehat{\mathbf{A}}-\mathbf{A}\|_{1} = \|\mathbf{D}_{\mathbf{Q}}(\widehat{\mathbf{B}}-\mathbf{B})\mathbf{D}_{\mathbf{K}}\|_{1} \leq 
\|\mathbf{D}_{\mathbf{Q}}\|_{1}
\|\widehat{\mathbf{B}}-\mathbf{B}\|_{1}
\|\mathbf{D}_{\mathbf{K}}\|_{1} \leq \delta g^{*}h^{*}
\end{equation}
Taking $\delta = \frac{\epsilon}{g^{*}h^{*}}$ completes the proof.
\end{proof}

\section{Hyperparameters}
\label{appendix:hyperparameters}
This optimal setting (including comparisons to approximate softmax) we use for the Performer is specified in the Generalized Attention (Subsection \ref{subsec:generalized_default}), and \textbf{unless specifically mentioned (e.g. using name "Performer-SOFTMAX"), "Performer" refers to using this generalized attention setting.}

\subsection{Training Hyperparameters} All Performer + Transformer runs used $0.5$ grad clip, $0.1$ weight decay, $0.1$ dropout, $10^{-3}$ fixed learning rate with Adam hyperparameters $(\beta_{1} = 0.9, \beta_{2} = 0.98, \epsilon = 10^{-9})$, with batch size maximized (until TPU memory overload) for a specific model. For the Reformer, we used the same hyperparameters as mentioned for protein experiments, without gradient clipping, while using the defaults\footnote{\url{https://github.com/google/trax/blob/master/trax/supervised/configs/reformer_imagenet64.gin}} (which instead use learning rate decay) for ImageNet-64. In both cases, the Reformer used the same default LSH attention parameters.

All 36-layer protein experiments used the same amount of compute (i.e. 16x16 TPU-v2, 8GB per chip). For concatenated experiments, 16x16 TPU-v2's were also used for the Performer, while 8x8's were used for the 1-3 layer $(d = 256)$ Transformer models (using 16x16 did not make a difference in accuracy).

\subsection{Approximate Softmax Attention Default Values} The optimal values, set to default parameters\footnote{\url{https://github.com/google-research/google-research/blob/master/performer/fast_self_attention/fast_self_attention.py\#L198}}, are: renormalize\_attention = True, numerical stabilizer = $10^{-6}$, number of features = 256, ortho\_features = True, ortho\_scaling = 0.0.  .

\subsection{Generalized Attention Default Values} \label{subsec:generalized_default} The optimal values, set to default parameters\footnote{\url{https://github.com/google-research/google-research/blob/master/performer/fast_self_attention/fast_self_attention.py\#L260}}, are: renormalize\_attention = True, numerical stabilizer = 0.0, number of features = 256, kernel = ReLU, kernel\_epsilon = $10^{-3}$. 

\clearpage

\section{Experimental Details for Protein Modeling Tasks}
\label{sec:protein_extended}
\subsection{TrEMBL Dataset}

\begin{table}[h]
\centering
\renewcommand{\arraystretch}{1.4}
\resizebox{0.9\textwidth}{!}{%
\begin{tabular}{ |c|c|c|c|c|c|c|c|c| } 
\hline
\multirow{2}{*}{\bf Dataset} & \multirow{2}{*}{\bf Set Name} & \multirow{2}{*}{\bf Count} & \multicolumn{5}{|c|}{\bf Length Statistics} \\
\cline{4-8}
 & &  & \textbf{Min} & \textbf{Max}& \textbf{Mean}& \textbf{STD} & \textbf{Median} \\
\hline
\multirow{4}{*}{TrEMBL}
& Train & 104,863,744 & 2 & 74,488 & 353.09 & 311.16 & 289.00 \\ 
& Valid & 102,400 & 7 & 11,274 & 353.62 & 307.42 & 289.00 \\ 
\cline{2-8}
& Test & 1,033,216 & 8 & 32,278 & 353.96 & 312.23 & 289.00 \\ 
& OOD & 29,696 & 24 & 4,208 & 330.96 & 269.86 & 200.00 \\ 
\hline
& & & & & & & \vspace{-4.5mm}\\
\hline
\multirow{2}{*}{\begin{tabular}{c}TrEMBL\\ (concat)\end{tabular}} & Train & 4,532,224 & \multirow{2}{*}{8,192} & \multirow{2}{*}{8,192} & \multirow{2}{*}{8,192} & \multirow{2}{*}{0} & \multirow{2}{*}{8,192}\\
& Valid & 4,096 & & & & & \\ 
\hline
\end{tabular}
}
\vspace{2mm}
\caption{Statistics for the TrEMBL single sequence and the long sequence task.}
\label{table-trembl-statistics}
\end{table}

We used the TrEMBL dataset\footnote{\url{https://www.uniprot.org/statistics/TrEMBL}}, which contains 139,394,261 sequences of which 106,030,080 are unique. While the training dataset appears smaller than the one used in Madani et al. \cite{progen}, we argue that it includes most of the relevant sequences. Specifically, the TrEMBL dataset consists of the subset of UniProtKB sequences that have been computationally analyzed but not manually curated, and accounts for $\approx99.5\%$ of the total number of sequences in the UniProtKB dataset\footnote{\url{https://www.uniprot.org/uniprot/}}.

Following the methodology described in Madani et al. \cite{progen}, we used both an OOD-Test set, where a selected subset of Pfam families are held-out for valuation, and an IID split, where the remaining protein sequences are split randomly into train, valid, and test tests. We held-out the following protein families (PF18369, PF04680, PF17988, PF12325, PF03272, PF03938, PF17724, PF10696, PF11968, PF04153, PF06173, PF12378, PF04420, PF10841, PF06917, PF03492, PF06905, PF15340, PF17055, PF05318), which resulted in 29,696 OOD sequences. We note that, due to deduplication and potential TrEMBL version mismatch, our OOD-Test set does not match exactly the one in Madani et al. \cite{progen}. We also note that this OOD-Test selection methodology does not guarantee that the evaluation sequences are within a minimum distance from the sequences used during training. In future work, we will include rigorous distance based splits.

The statistics for the resulting dataset splits are reported in Table \ref{table-trembl-statistics}. In the standard sequence modeling task, given the length statistics that are reported in the table, we clip single sequences to maximum length $L=1024$, which results in few sequences being truncated significantly.

In the long sequence task, the training and validation sets are obtained by concatenating the sequences, separated by an end-of-sequence token, and grouping the resulting chain into non-overlapping sequences of length $L=8192$.


\subsection{Empirical Baseline}

\begin{figure}[h]
  \centering
  \includegraphics[width=0.99\linewidth]{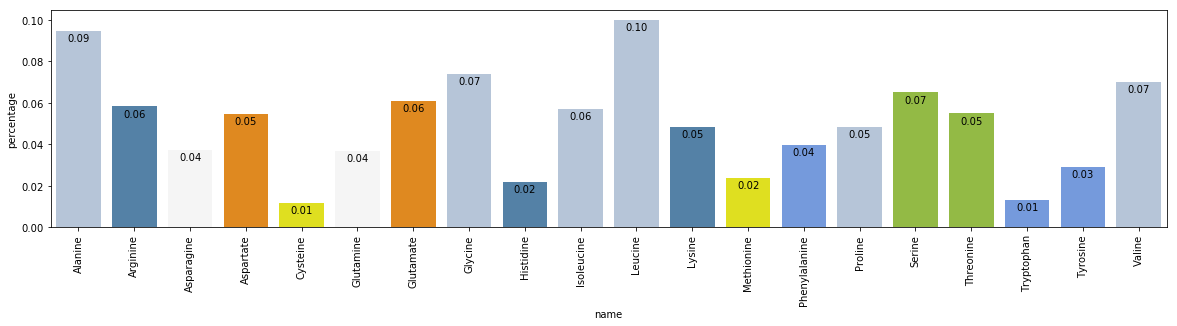}
  \caption{\small{Visualization of the estimated empirical distribution for the 20 standard amino acids, colored by their class. Note the consistency with the statistics on the TrEMBL web page.}}
  \vspace{2mm}
  \label{figure-empirical-baseline}
\end{figure}

A random baseline, with uniform probability across all the vocabulary tokens at every position, has accuracy $5\%$ (when including only the 20 standard amino acids) and $4\%$ (when also including the 5 anomalous amino acids \cite{uniprot2019uniprot}). However, the empirical frequencies of the various amino acids in our dataset may be far from uniform, so we also consider an \textit{empirical baseline} where the amino acid probabilities are proportional to their empirical frequencies in the training set.

Figure \ref{figure-empirical-baseline} shows the estimated empirical distribution. We use both the standard and anomalous amino acids, and we crop sequences to length 1024 to match the data processing performed for the Transformer models. The figure shows only the 20 standard amino acids, colored by their class, for comparison with the visualization on the TrEMBL web page\footnote{\url{https://www.uniprot.org/statistics/TrEMBL}}.

\subsection{Tabular Results}

Table \ref{table-trembl} contains the results on the single protein sequence modeling task ($L=1024$). We report the following evaluation metrics:
\begin{enumerate}
\item \textbf{Accuracy}: For unidirectional models, we measure the accuracy on next-token prediction, averaged across all sequence positions in the dataset. For bidirectional models, we mask each token with $15\%$ probability and measure accuracy across the masked positions.
\item \textbf{Perplexity}: For unidirectional models, we measure perplexity across all sequence positions in the dataset. For bidirectional models, similar to the accuracy case, we measure perplexity across the masked positions. 
\end{enumerate}

\begin{table}[h]
\centering
\renewcommand{\arraystretch}{1.4}
\resizebox{0.9\textwidth}{!}{%
\begin{tabular}{ |c|c|c|c|c| } 
\hline
\textbf{Model Type} & \textbf{Set Name} & \textbf{Model} & \textbf{Accuracy} & \textbf{Perplexity} \\
\hline
\multirow{6}{*}{UNI} & \multirow{3}{*}{Test} & Empirical Baseline & 9.92 & 17.80 \\ 
& & Transformer & 30.80 & 9.37\\ 
& & Performer (generalized) & 31.58 & 9.17 \\ 
\cline{2-5}
& \multirow{3}{*}{OOD} & Empirical Baseline & ~9.07 & ~17.93 \\ 
& & Transformer & 19.70 & 13.20 \\ 
& & Performer (generalized) & 18.44 & 13.63 \\
\hline
& & & &\vspace{-4.5mm}\\
\hline
\multirow{6}{*}{BID} & \multirow{3}{*}{Test} & Transformer & 33.32 & 9.22 \\ 
& & Performer (generalized) & 36.09 &  8.36 \\ 
& & Performer (softmax) & 33.00 &  9.24 \\ 
\cline{2-5}
& \multirow{3}{*}{OOD} & Transformer & 25.07 & 12.09 \\ 
& & Performer (generalized) & 24.10 & 12.26 \\ 
& & Performer (softmax) & 23.48 & 12.41 \\ 
\hline
\end{tabular}
}
\vspace{2mm}
\caption{Results on single protein sequence modeling ($L=1024$). We note that the empirical baseline results are applicable to both the unidirectional (UNI) and bidirectional (BID) models.}\label{table-trembl}
\end{table}

\subsection{Attention Matrix Illustration}

In this section we illustrate the attention matrices produced by a Performer model. We focus on the bidirectional case and choose one Performer model trained on the standard single-sequence TrEMBL task for over 500K steps. The same analysis can be applied to unidirectional Performers as well.

We note that while the Transformer model instantiates the attention matrix in order to compute the attention output that incorporates the (queries $Q$, keys $K$, values $V$) triplet (see Eq.~\ref{eq:attnorm} in the main paper), the FAVOR mechanism returns the attention output directly (see Algorithm~\ref{alg:1}). To account for this discrepancy, we extract the attention matrices by applying each attention mechanism twice: once on each original $(Q, K, V)$ triple to obtain the attention output, and once on a modified $(Q, K, V^\circ)$ triple, where $V^\circ$ contains one-hot indicators for each position index, to obtain the attention matrix. The choice of $V^\circ$ ensures that the dimension of the attention output is equal to the sequence length, and that a non-zero output on a dimension $i$ can only arise from a non-zero attention weight to the $i^{th}$ sequence position. Indeed, in the Transformer case, when comparing the output of this procedure with the instantiated attention matrix, the outputs match.

\begin{figure}[h]
  \centering
  \includegraphics[width=0.99\textwidth]{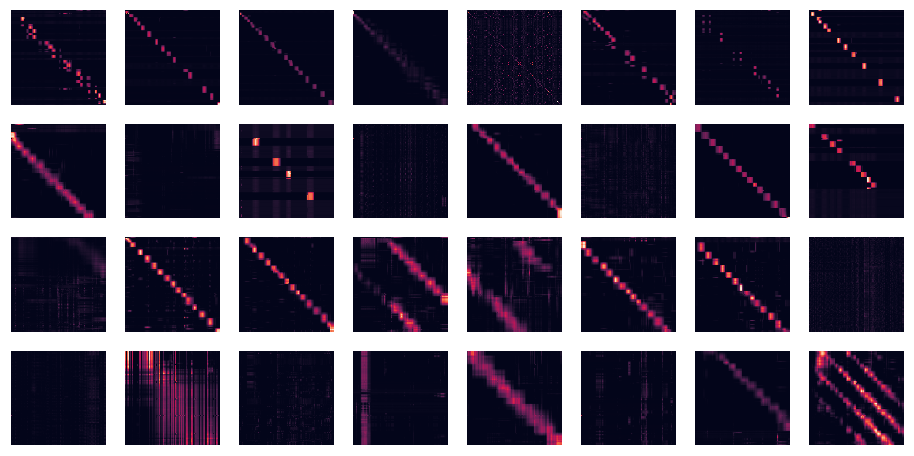}
  \vspace{2mm}
  \caption{\small{We show the attention matrices for the first 4 layers and all 8 heads (each row is a layer, each column is head index, each cell contains the attention matrix across the entire BPT1\_BOVIN protein sequence). Note that many heads show a \textit{diagonal} pattern, where each node attends to its neighbors, and some heads show a \textit{vertical} pattern, where each head attends to the same fixed positions.}}
  \label{fig:attention_matrices}
\end{figure}

\begin{figure}[h]
  \centering
  \includegraphics[width=0.49\textwidth]{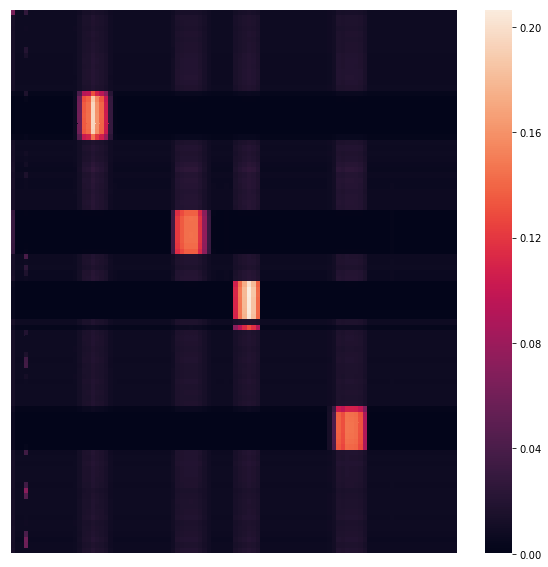}
  \includegraphics[width=0.49\textwidth]{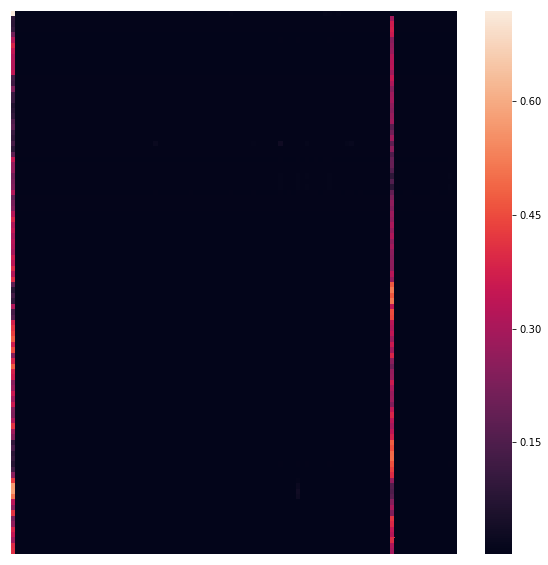}
  \vspace{2mm}
  \caption{\small{We illustrate in more detail two attention heads. The sub-figures correspond respectively to: \textbf{(1)} Head 1-2 (second layer, third head), \textbf{(2)} Head 4-1 (fifth layer, second head). Note the block attention in Head 1-2 and the vertical attention (to the start token (`M') and the 85th token (`C')) in Head 4-1.}}
  \label{fig:attention_heads}
\end{figure}

\begin{figure}[h]
    \centering
  \includegraphics[height = 0.6\linewidth, width=0.9\linewidth]{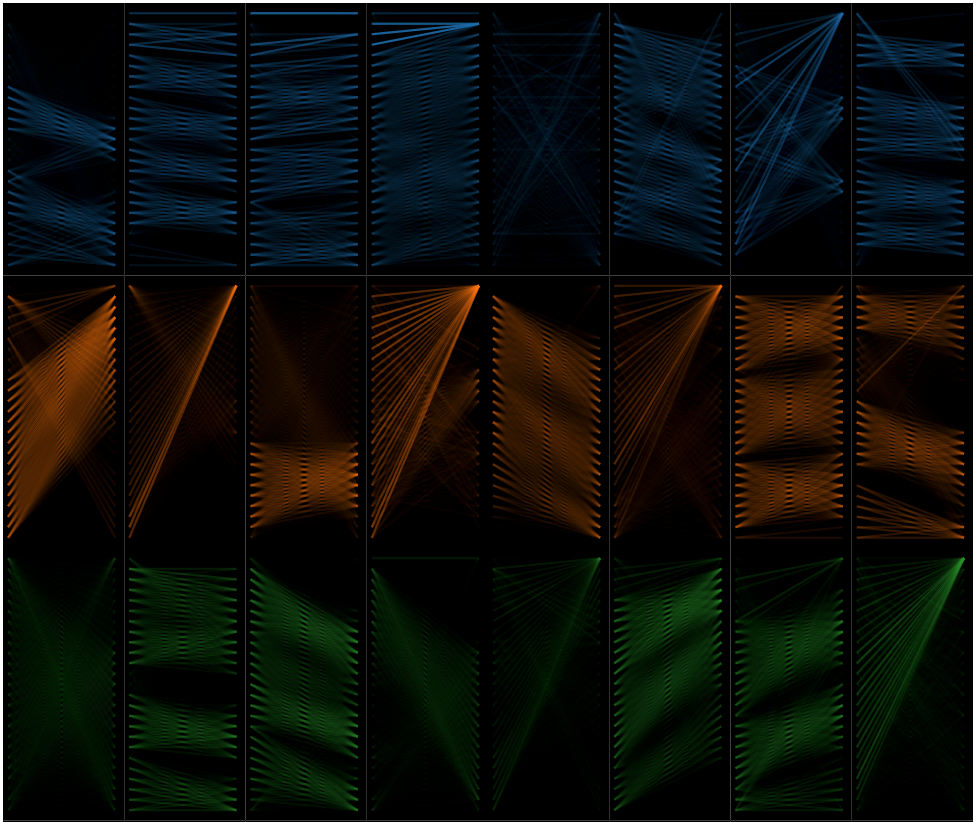}
  \vspace{2mm}
  \caption{\small{We highlight the attention patterns by restricting our attention to the first 25 tokens (note that we do not renormalize the attention to these tokens). The illustration is based on Vig et al. \protect\cite{vig2019multiscale, analyzing_attention}. Note that, similar to prior work on protein Transformers \protect\cite{progen}, the attention matrices include both local and global patterns.}}
  \label{fig:model_view}
\end{figure}

\textbf{Attention matrix example.} We start by visualizing the attention matrix for an individual protein sequence. We use the BPT1\_BOVIN protein sequence\footnote{\url{https://www.uniprot.org/uniprot/P00974}}, one of the most extensively studied globular proteins, which contains 100 amino acids. In Figure \ref{fig:attention_matrices}, we show the attention matrices for the first 4 layers. Note that many heads show a \textit{diagonal} pattern, where each node attends to its neighbors, and some heads show a \textit{vertical} pattern, where each head attends to the same fixed positions. These patterns are consistent with the patterns found in Transformer models trained on natural language \cite{kovaleva2019revealing}. In Figure \ref{fig:model_view} we highlight these attention patterns by focusing on the first 25 tokens, and in Figure \ref{fig:attention_heads}, we illustrate in more detail two attention heads.

\begin{figure}[h]
  \centering
  \includegraphics[width=0.49\textwidth]{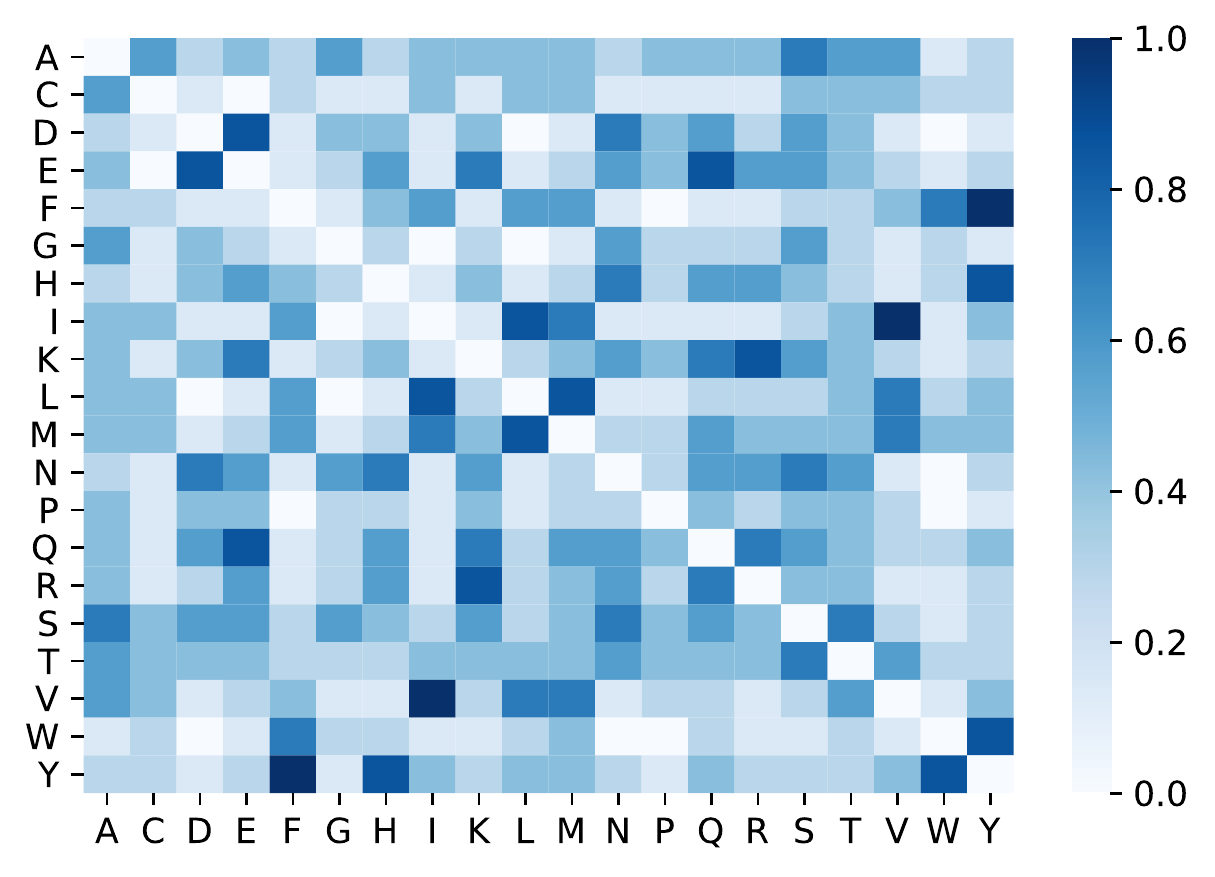}
  \includegraphics[width=0.49\textwidth]{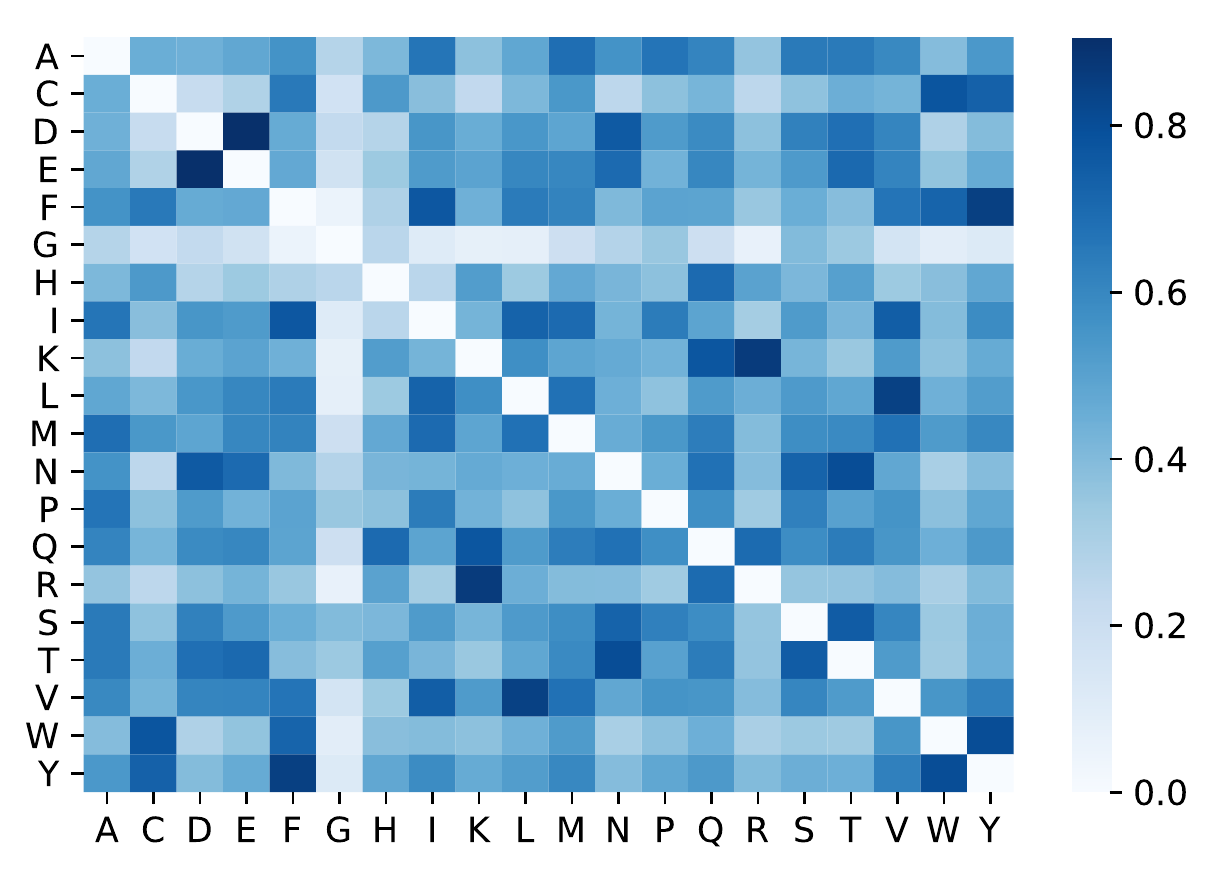}
  \vspace{1mm}
  \caption{\small{Amino acid similarity matrix estimated from attention matrices aggregated across a small subset of sequences, as described in Vig et al. \protect\cite{bertology}. The sub-figures correspond respectively to: \textbf{(1)} the normalized BLOSUM matrix, \textbf{(2)} the amino acid similarity estimated via a trained Performer model. Note that the Performer recognises highly similar amino acid pairs such as (D, E) and (F, Y).}}
  \label{figure:amino-acid-similarity}
\end{figure}

\textbf{Amino acid similarity.} Furthermore, we analyze the amino-acid similarity matrix estimated from the attention matrices produced by the Performer model, as described in Vig et al. \cite{bertology}. We aggregate the attention matrix across 800 sequences. The resulting similarity matrix is illustrated in Figure \ref{figure:amino-acid-similarity}. Note that the Performer recognises highly similar amino acid pairs such as (D, E) and (F, Y).

\section{Extended approximation results}
\label{appendix:extended_approx}

\subsection{Backwards Compatibility}
Although mentioned previously (Sec. \ref{subsec:approx_error_compatibility}) that the Performer with additional finetuning is backwards compatible with the Transformer, we demonstrate below in Fig. \ref{fig:appendix_approx} that error propagation due to non-attention components of the Transformer is one of the primary reasons that pretrained Transformer weights cannot be immediately used for inference on the corresponding Performer.
\begin{figure}[h]
  \centering
  \includegraphics[width=0.75\textwidth]{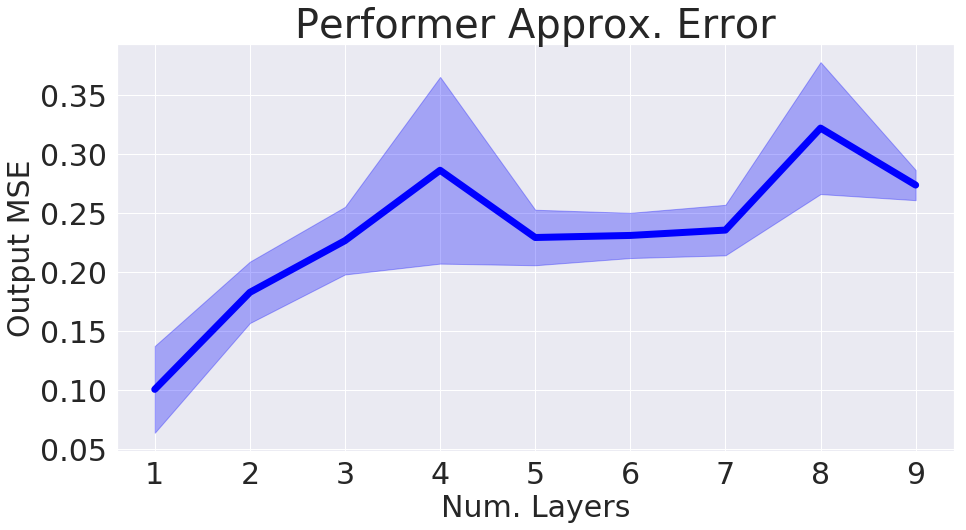}
  \caption{Output approximation errors between a vanilla Transformer and a Performer (with orthogonal features) for varying numbers of layers.}
  \label{fig:appendix_approx}
\end{figure}

\subsection{Generalized Attention}

\begin{figure}[h]
  \centering
  \includegraphics[width=0.99\textwidth]{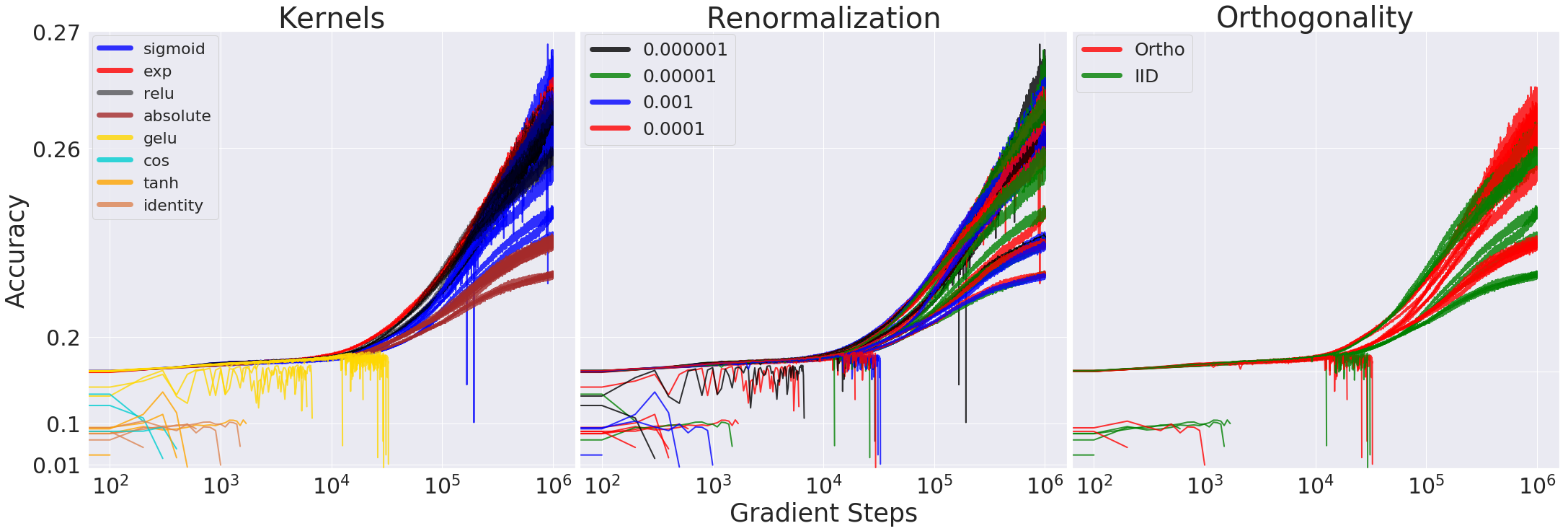}
  \caption{To emphasize the highest accuracy runs but also show the NaN issues with certain kernels which caused runs to stop early, we set both x and y axes to be log-scale. We tested kernels defined by different functions $f$ (see: Sec. \ref{sec:generalized}): sigmoid, exponential, ReLU, absolute, gelu, cosine (original softmax approximation), tanh, and identity. All training runs were performed on 2x2 TPU-v2's, 128 batch size per device.}
  \label{fig:attention_comparisons_2x2}
\end{figure}

\begin{figure}[h]
  \centering
  \includegraphics[width=0.99\textwidth]{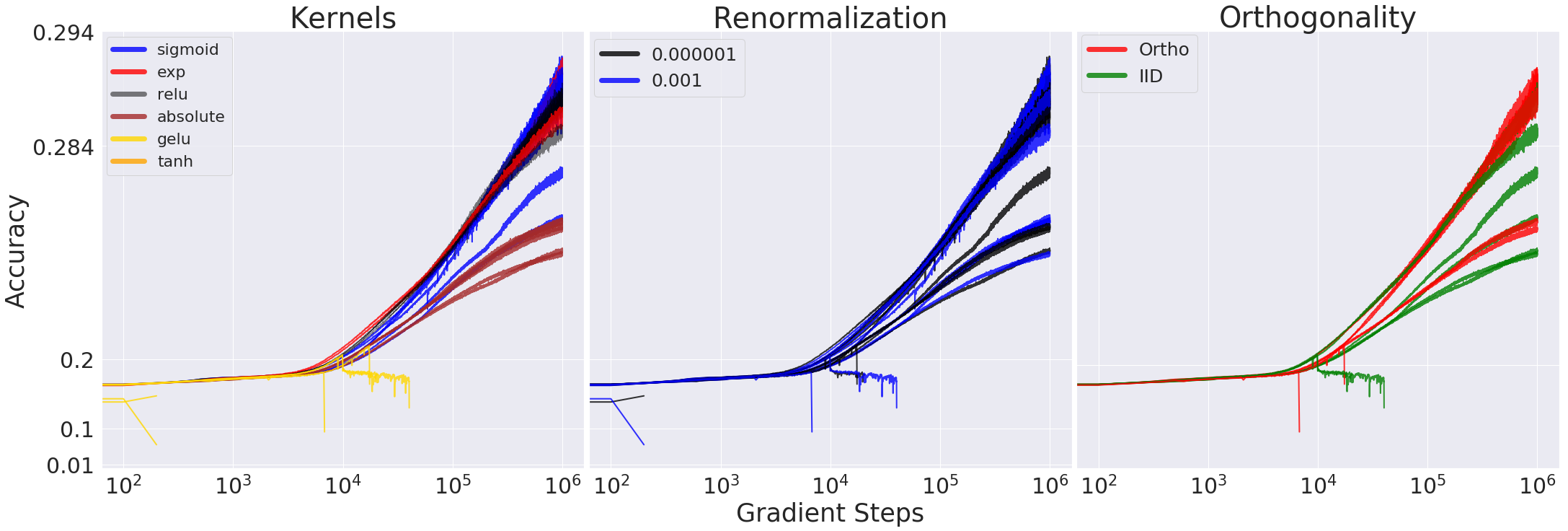}
  \caption{We also performed a similar setup as Fig. \ref{fig:attention_comparisons_2x2} for 4x4 TPU-v2's.}
  \label{fig:attention_comparisons_4x4}
\end{figure}

We investigated Generalized Attention mechanisms (mentioned in Sec. \ref{sec:generalized}) on TrEMBL when $L=512$ for various kernel functions. This is similar to \cite{tsai2019transformer} which also experiments with various attention kernels for natural language. Using hyperparameter sweeps across multiple variables in FAVOR, we compared several kernels and also renormalization on/off (Fig. \ref{fig:attention_comparisons_2x2} and Fig. \ref{fig:attention_comparisons_4x4}), where $\mathrm{Renormalize}$ corresponds to applying $\mathbf{D}^{-1}$ operator in attention, as for the standard mechanism, though we noticed that disabling it does not necessarily hurt accuracy) to produce the best training configuration for the Performer. We note that the effective batch size slightly affects the rankings (as shown by the difference between 2x2 and 4x4 TPU runs) - we by default use the generalized ReLU kernel with other default hyperparameters shown in Appendix \ref{appendix:hyperparameters}, as we observed that they are empirically optimal for large batch size runs (i.e. 8x8 or 16x16 TPU's).

\clearpage 

\section{Extended computation costs}

\begin{figure}[h]
  \includegraphics[width=1.0\textwidth]{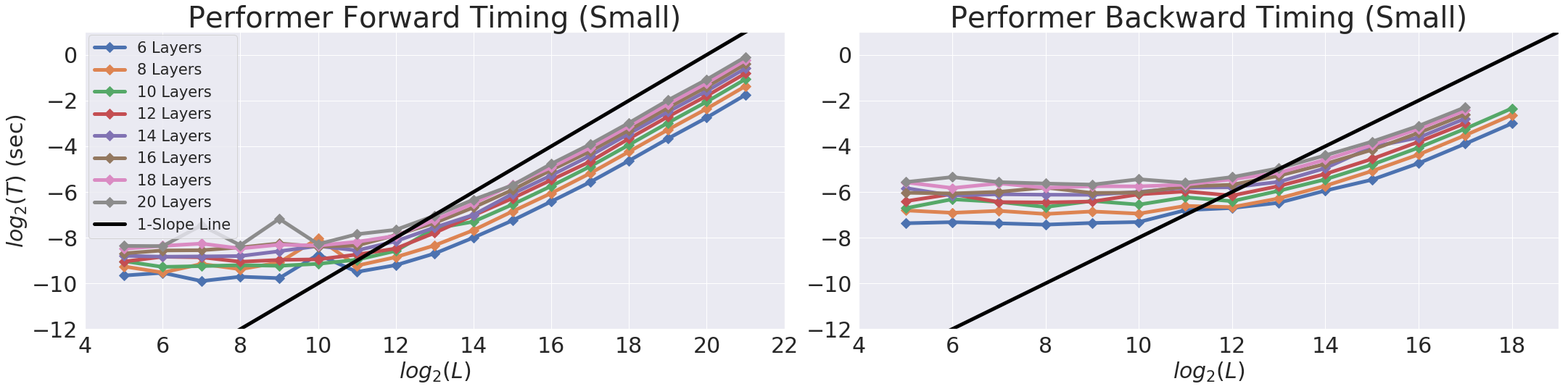}
  \includegraphics[width=1.0\textwidth]{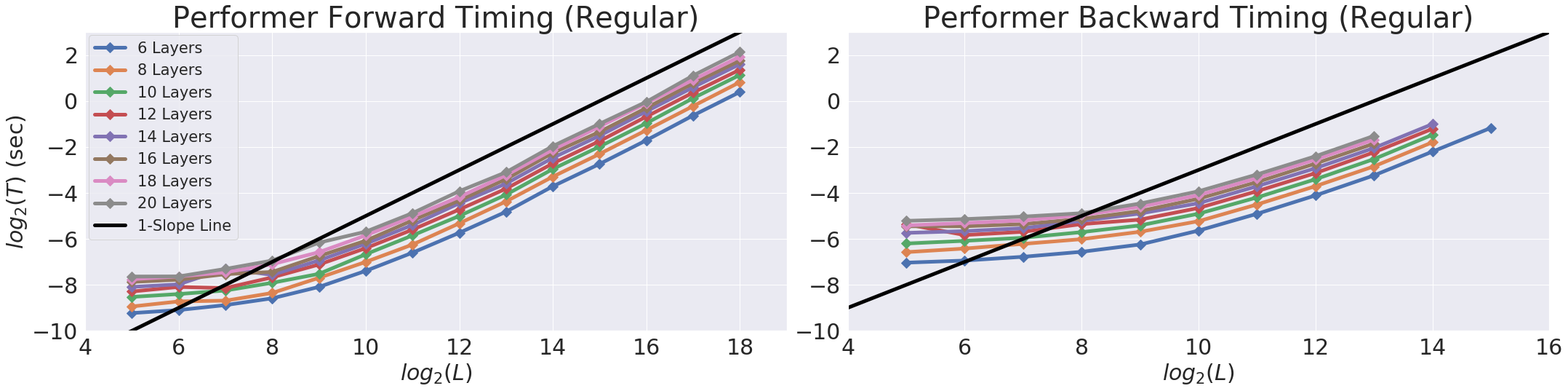}
  \includegraphics[width=1.0\textwidth]{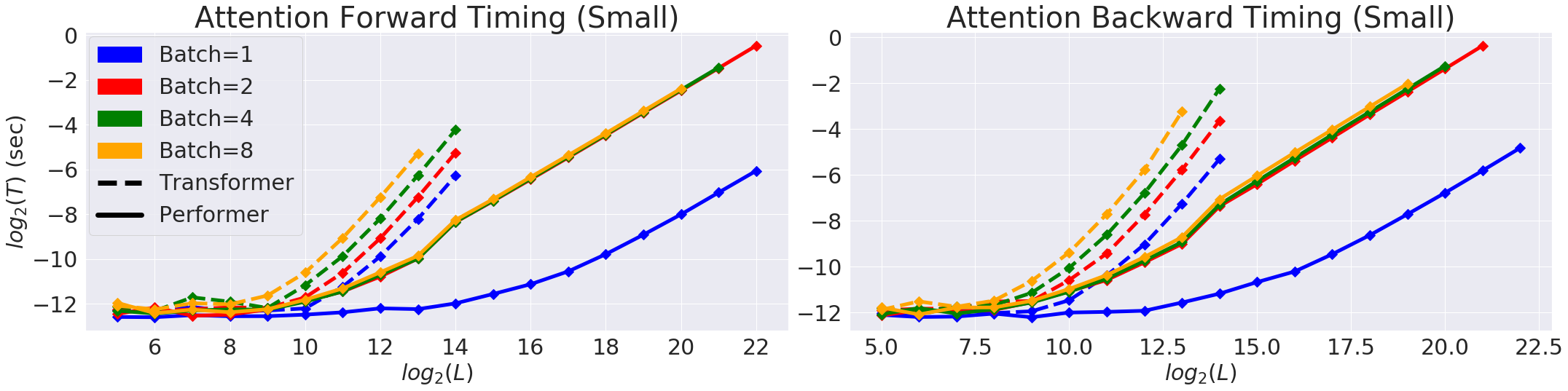}
  \includegraphics[width=1.0\textwidth]{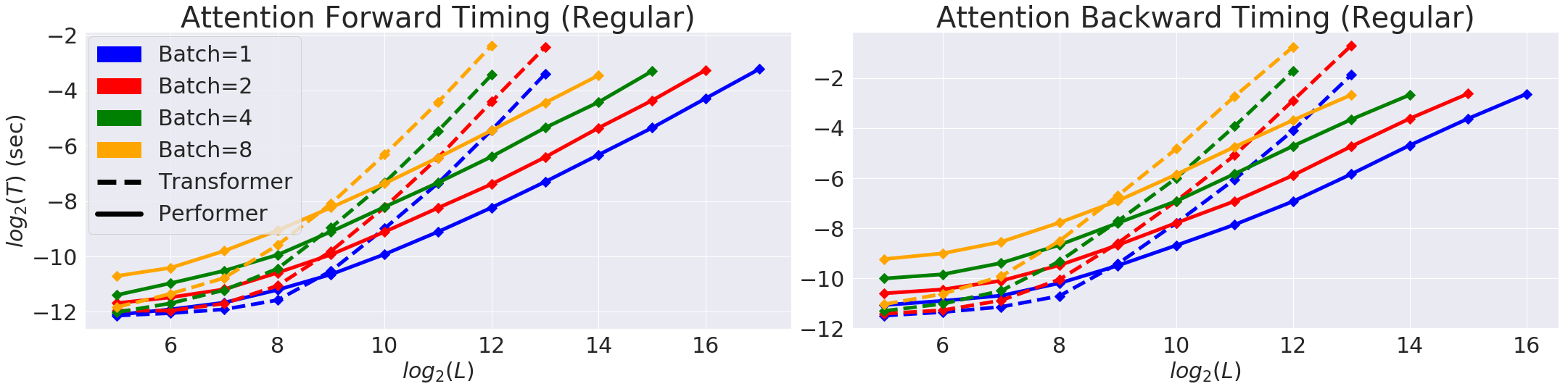}
  
  \caption{Captions (1), (2) for each 2x2 subfigure mentioned above.}
  \label{fig:appendix_runtime_1}
\end{figure}

\begin{figure}[h]
  \includegraphics[width=1.0\textwidth]{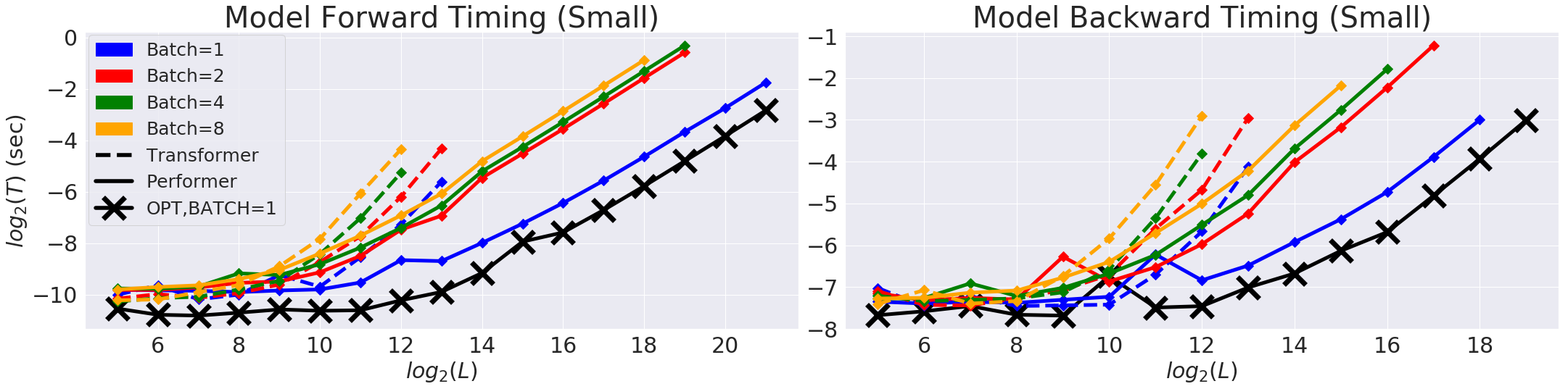}
  \includegraphics[width=1.0\textwidth]{img/model_regular_opt.png}

  \caption{Caption (3) for this 2x2 subfigure mentioned above.}
  \label{fig:appendix_runtime_2}
\end{figure}

In this subsection, we empirically measure computational costs in terms wall clock time on forward and backward passes for three scenarios in Fig. \ref{fig:appendix_runtime_1}, \ref{fig:appendix_runtime_2}:
\begin{enumerate}
\item Performer, with varying number of layers. We show that our method can scale up to (but not necessarily limited to) even 20 layers.
\item Attention time complexities when comparing standard attention (from Transformer) and FAVOR (from Performer). Note that the maximum memory size here is not reflective of the maximum memory size in an actual model (shown below), as this benchmark requires computing explicit tensors (causing memory increases) in Jax, while a model does not.
\item Time complexities when comparing the Transformer and Performer models. "X" (OPT) denotes the maximum possible speedup achievable, when attention simply returns the $\mathbf{V}$-vector, showing that the Performer is nearly optimal. We see that the maximum possible power of 2 length allowed on a V100 GPU (16GB) is $2^{15} = 32768$ using regular dimensions.
\end{enumerate}

Since some of the computational bottleneck in the Transformer may originate from the extra feed-forward layers \cite{reformer}, we also benchmark the ``Small" version, i.e. $(n_{heads}, n_{layers}, d_{ff}, d) = (1,6,64,64)$ as well, when the attention component is the dominant source of computation and memory. We remind the reader that the ``Regular" version consists of  $(n_{heads}, n_{layers}, d_{ff}, d) = (8,6,2048,512)$. 

\label{appendix:computation_costs_bidirectional}

\clearpage

%% file: main.bbl
\begin{thebibliography}{10}

\bibitem{image_transformer}
I.~Bello, B.~Zoph, A.~Vaswani, J.~Shlens, and Q.~V. Le.
\newblock Attention augmented convolutional networks.
\newblock {\em CoRR}, abs/1904.09925, 2019.

\bibitem{longformer}
I.~Beltagy, M.~E. Peters, and A.~Cohan.
\newblock Longformer: The long-document transformer.
\newblock {\em CoRR}, abs/2004.05150, 2020.

\bibitem{bitbol2016inferring}
A.-F. Bitbol, R.~S. Dwyer, L.~J. Colwell, and N.~S. Wingreen.
\newblock Inferring interaction partners from protein sequences.
\newblock {\em Proceedings of the National Academy of Sciences},
  113(43):12180--12185, 2016.

\bibitem{imputer}
W.~Chan, C.~Saharia, G.~E. Hinton, M.~Norouzi, and N.~Jaitly.
\newblock Imputer: Sequence modelling via imputation and dynamic programming.
\newblock {\em CoRR}, abs/2002.08926, 2020.

\bibitem{chelba}
C.~Chelba, M.~X. Chen, A.~Bapna, and N.~Shazeer.
\newblock Faster transformer decoding: N-gram masked self-attention.
\newblock {\em CoRR}, abs/2001.04589, 2020.

\bibitem{lm1b}
C.~Chelba, T.~Mikolov, M.~Schuster, Q.~Ge, T.~Brants, P.~Koehn, and
  T.~Robinson.
\newblock One billion word benchmark for measuring progress in statistical
  language modeling.
\newblock In {\em {INTERSPEECH} 2014, 15th Annual Conference of the
  International Speech Communication Association, Singapore, September 14-18,
  2014}, pages 2635--2639, 2014.

\bibitem{nmt}
M.~X. Chen, O.~Firat, A.~Bapna, M.~Johnson, W.~Macherey, G.~F. Foster,
  L.~Jones, M.~Schuster, N.~Shazeer, N.~Parmar, A.~Vaswani, J.~Uszkoreit,
  L.~Kaiser, Z.~Chen, Y.~Wu, and M.~Hughes.
\newblock The best of both worlds: Combining recent advances in neural machine
  translation.
\newblock In I.~Gurevych and Y.~Miyao, editors, {\em Proceedings of the 56th
  Annual Meeting of the Association for Computational Linguistics, {ACL} 2018,
  Melbourne, Australia, July 15-20, 2018, Volume 1: Long Papers}, pages 76--86.
  Association for Computational Linguistics, 2018.

\bibitem{sparsetr}
R.~Child, S.~Gray, A.~Radford, and I.~Sutskever.
\newblock Generating long sequences with sparse transformers.
\newblock {\em CoRR}, abs/1904.10509, 2019.

\bibitem{psrnn}
K.~Choromanski, C.~Downey, and B.~Boots.
\newblock Initialization matters: Orthogonal predictive state recurrent neural
  networks.
\newblock In {\em 6th International Conference on Learning Representations,
  {ICLR} 2018, Vancouver, BC, Canada, April 30 - May 3, 2018, Conference Track
  Proceedings}. OpenReview.net, 2018.

\bibitem{kama}
K.~Choromanski, A.~Pacchiano, J.~Pennington, and Y.~Tang.
\newblock {KAMA-NN}s: Low-dimensional rotation based neural networks.
\newblock In K.~Chaudhuri and M.~Sugiyama, editors, {\em The 22nd International
  Conference on Artificial Intelligence and Statistics, {AISTATS} 2019, 16-18
  April 2019, Naha, Okinawa, Japan}, volume~89 of {\em Proceedings of Machine
  Learning Research}, pages 236--245. {PMLR}, 2019.

\bibitem{uni}
K.~Choromanski, M.~Rowland, W.~Chen, and A.~Weller.
\newblock Unifying orthogonal {Monte Carlo} methods.
\newblock In K.~Chaudhuri and R.~Salakhutdinov, editors, {\em Proceedings of
  the 36th International Conference on Machine Learning, {ICML} 2019, 9-15 June
  2019, Long Beach, California, {USA}}, volume~97 of {\em Proceedings of
  Machine Learning Research}, pages 1203--1212. {PMLR}, 2019.

\bibitem{geom}
K.~Choromanski, M.~Rowland, T.~Sarl{\'{o}}s, V.~Sindhwani, R.~E. Turner, and
  A.~Weller.
\newblock The geometry of random features.
\newblock In A.~J. Storkey and F.~P{\'{e}}rez{-}Cruz, editors, {\em
  International Conference on Artificial Intelligence and Statistics, {AISTATS}
  2018, 9-11 April 2018, Playa Blanca, Lanzarote, Canary Islands, Spain},
  volume~84 of {\em Proceedings of Machine Learning Research}, pages 1--9.
  {PMLR}, 2018.

\bibitem{unreas}
K.~M. Choromanski, M.~Rowland, and A.~Weller.
\newblock The unreasonable effectiveness of structured random orthogonal
  embeddings.
\newblock In I.~Guyon, U.~von Luxburg, S.~Bengio, H.~M. Wallach, R.~Fergus,
  S.~V.~N. Vishwanathan, and R.~Garnett, editors, {\em Advances in Neural
  Information Processing Systems 30: Annual Conference on Neural Information
  Processing Systems 2017, 4-9 December 2017, Long Beach, CA, {USA}}, pages
  219--228, 2017.

\bibitem{cong2019protein}
Q.~Cong, I.~Anishchenko, S.~Ovchinnikov, and D.~Baker.
\newblock Protein interaction networks revealed by proteome coevolution.
\newblock {\em Science}, 365(6449):185--189, 2019.

\bibitem{uniprot2019uniprot}
U.~Consortium.
\newblock Uniprot: a worldwide hub of protein knowledge.
\newblock {\em Nucleic acids research}, 47(D1):D506--D515, 2019.

\bibitem{cormen}
T.~H. Cormen, C.~E. Leiserson, R.~L. Rivest, and C.~Stein.
\newblock {\em Introduction to Algorithms, 3rd Edition}.
\newblock {MIT} Press, 2009.

\bibitem{transformerxl}
Z.~Dai*, Z.~Yang*, Y.~Yang, W.~W. Cohen, J.~Carbonell, Q.~V. Le, and
  R.~Salakhutdinov.
\newblock Transformer-{XL}: Language modeling with longer-term dependency,
  2019.

\bibitem{universal_t}
M.~Dehghani, S.~Gouws, O.~Vinyals, J.~Uszkoreit, and L.~Kaiser.
\newblock Universal transformers.
\newblock In {\em 7th International Conference on Learning Representations,
  {ICLR} 2019, New Orleans, LA, USA, May 6-9, 2019}. OpenReview.net, 2019.

\bibitem{bert}
J.~Devlin, M.~Chang, K.~Lee, and K.~Toutanova.
\newblock {BERT:} pre-training of deep bidirectional transformers for language
  understanding.
\newblock {\em CoRR}, abs/1810.04805, 2018.

\bibitem{du2020energy}
Y.~Du, J.~Meier, J.~Ma, R.~Fergus, and A.~Rives.
\newblock Energy-based models for atomic-resolution protein conformations.
\newblock {\em arXiv preprint arXiv:2004.13167}, 2020.

\bibitem{elnaggar2019end}
A.~Elnaggar, M.~Heinzinger, C.~Dallago, and B.~Rost.
\newblock End-to-end multitask learning, from protein language to protein
  features without alignments.
\newblock {\em bioRxiv}, page 864405, 2019.

\bibitem{jax}
R.~Frostig, M.~Johnson, and C.~Leary.
\newblock Compiling machine learning programs via high-level tracing.
\newblock In {\em Conference on Machine Learning and Systems 2018}, 2018.

\bibitem{attention_cvpr}
J.~Fu, J.~Liu, H.~Tian, Y.~Li, Y.~Bao, Z.~Fang, and H.~Lu.
\newblock Dual attention network for scene segmentation.
\newblock In {\em {IEEE} Conference on Computer Vision and Pattern Recognition,
  {CVPR} 2019, Long Beach, CA, USA, June 16-20, 2019}, pages 3146--3154, 2019.

\bibitem{conformer}
A.~Gulati, J.~Qin, C.-C. Chiu, N.~Parmar, Y.~Zhang, J.~Yu, W.~Han, S.~Wang,
  Z.~Zhang, Y.~Wu, and R.~Pang.
\newblock Conformer: Convolution-augmented transformer for speech recognition,
  2020.

\bibitem{gulrajani}
I.~Gulrajani, F.~Ahmed, M.~Arjovsky, V.~Dumoulin, and A.~C. Courville.
\newblock Improved training of {Wasserstein GANs}.
\newblock In I.~Guyon, U.~von Luxburg, S.~Bengio, H.~M. Wallach, R.~Fergus,
  S.~V.~N. Vishwanathan, and R.~Garnett, editors, {\em Advances in Neural
  Information Processing Systems 30: Annual Conference on Neural Information
  Processing Systems 2017, 4-9 December 2017, Long Beach, CA, {USA}}, pages
  5767--5777, 2017.

\bibitem{hopf2012three}
T.~A. Hopf, L.~J. Colwell, R.~Sheridan, B.~Rost, C.~Sander, and D.~S. Marks.
\newblock Three-dimensional structures of membrane proteins from genomic
  sequencing.
\newblock {\em Cell}, 149(7):1607--1621, 2012.

\bibitem{simon}
C.~A. Huang, A.~Vaswani, J.~Uszkoreit, I.~Simon, C.~Hawthorne, N.~Shazeer,
  A.~M. Dai, M.~D. Hoffman, M.~Dinculescu, and D.~Eck.
\newblock Music transformer: Generating music with long-term structure.
\newblock In {\em 7th International Conference on Learning Representations,
  {ICLR} 2019, New Orleans, LA, USA, May 6-9, 2019}. OpenReview.net, 2019.

\bibitem{ingraham2019generative}
J.~Ingraham, V.~Garg, R.~Barzilay, and T.~Jaakkola.
\newblock Generative models for graph-based protein design.
\newblock In {\em Advances in Neural Information Processing Systems}, pages
  15794--15805, 2019.

\bibitem{reformer}
N.~Kitaev, L.~Kaiser, and A.~Levskaya.
\newblock Reformer: The efficient transformer.
\newblock In {\em 8th International Conference on Learning Representations,
  {ICLR} 2020, Addis Ababa, Ethiopia, April 26-30, 2020}. OpenReview.net, 2020.

\bibitem{kovaleva2019revealing}
O.~Kovaleva, A.~Romanov, A.~Rogers, and A.~Rumshisky.
\newblock Revealing the dark secrets of bert.
\newblock {\em arXiv preprint arXiv:1908.08593}, 2019.

\bibitem{cumsum}
R.~E. Ladner and M.~J. Fischer.
\newblock Parallel prefix computation.
\newblock {\em J. ACM}, 27(4):831–838, Oct. 1980.

\bibitem{Lin2020DemystifyingOM}
H.~Lin, H.~Chen, T.~Zhang, C.~Laroche, and K.~Choromanski.
\newblock Demystifying orthogonal {Monte Carlo} and beyond.
\newblock {\em CoRR}, abs/2005.13590, 2020.

\bibitem{luo}
H.~Luo, S.~Zhang, M.~Lei, and L.~Xie.
\newblock Simplified self-attention for transformer-based end-to-end speech
  recognition.
\newblock {\em CoRR}, abs/2005.10463, 2020.

\bibitem{progen}
A.~Madani, B.~McCann, N.~Naik, N.~S. Keskar, N.~Anand, R.~R. Eguchi, P.~Huang,
  and R.~Socher.
\newblock Progen: Language modeling for protein generation.
\newblock {\em CoRR}, abs/2004.03497, 2020.

\bibitem{marcandalli2019induction}
J.~Marcandalli, B.~Fiala, S.~Ols, M.~Perotti, W.~de~van~der Schueren,
  J.~Snijder, E.~Hodge, M.~Benhaim, R.~Ravichandran, L.~Carter, et~al.
\newblock Induction of potent neutralizing antibody responses by a designed
  protein nanoparticle vaccine for respiratory syncytial virus.
\newblock {\em Cell}, 176(6):1420--1431, 2019.

\bibitem{ovchinnikov2014robust}
S.~Ovchinnikov, H.~Kamisetty, and D.~Baker.
\newblock Robust and accurate prediction of residue--residue interactions
  across protein interfaces using evolutionary information.
\newblock {\em Elife}, 3:e02030, 2014.

\bibitem{parmar}
N.~Parmar, A.~Vaswani, J.~Uszkoreit, L.~Kaiser, N.~Shazeer, A.~Ku, and D.~Tran.
\newblock Image transformer.
\newblock In J.~G. Dy and A.~Krause, editors, {\em Proceedings of the 35th
  International Conference on Machine Learning, {ICML} 2018,
  Stockholmsm{\"{a}}ssan, Stockholm, Sweden, July 10-15, 2018}, volume~80 of
  {\em Proceedings of Machine Learning Research}, pages 4052--4061. {PMLR},
  2018.

\bibitem{compr}
J.~W. Rae, A.~Potapenko, S.~M. Jayakumar, C.~Hillier, and T.~P. Lillicrap.
\newblock Compressive transformers for long-range sequence modelling.
\newblock In {\em International Conference on Learning Representations}, 2020.

\bibitem{fourierapprox}
A.~Rahimi and B.~Recht.
\newblock Random features for large-scale kernel machines.
\newblock In J.~C. Platt, D.~Koller, Y.~Singer, and S.~T. Roweis, editors, {\em
  Advances in Neural Information Processing Systems 20, Proceedings of the
  Twenty-First Annual Conference on Neural Information Processing Systems,
  Vancouver, British Columbia, Canada, December 3-6, 2007}, pages 1177--1184.
  Curran Associates, Inc., 2007.

\bibitem{rives}
A.~Rives, S.~Goyal, J.~Meier, D.~Guo, M.~Ott, C.~Zitnick, J.~Ma, and R.~Fergus.
\newblock Biological structure and function emerge from scaling unsupervised
  learning to 250 million protein sequences.
\newblock {\em bioArxiv}, 04 2019.

\bibitem{hron}
M.~Rowland, J.~Hron, Y.~Tang, K.~Choromanski, T.~Sarl{\'{o}}s, and A.~Weller.
\newblock Orthogonal estimation of {W}asserstein distances.
\newblock In K.~Chaudhuri and M.~Sugiyama, editors, {\em The 22nd International
  Conference on Artificial Intelligence and Statistics, {AISTATS} 2019, 16-18
  April 2019, Naha, Okinawa, Japan}, volume~89 of {\em Proceedings of Machine
  Learning Research}, pages 186--195. {PMLR}, 2019.

\bibitem{routing_t}
A.~Roy, M.~Saffar, A.~Vaswani, and D.~Grangier.
\newblock Efficient content-based sparse attention with routing transformers.
\newblock {\em CoRR}, abs/2003.05997, 2020.

\bibitem{co2}
E.~Strubell, A.~Ganesh, and A.~McCallum.
\newblock Energy and policy considerations for deep learning in {NLP}.
\newblock {\em CoRR}, abs/1906.02243, 2019.

\bibitem{tang}
Y.~Tang, D.~Nguyen, and D.~Ha.
\newblock Neuroevolution of self-interpretable agents.
\newblock {\em CoRR}, abs/2003.08165, 2020.

\bibitem{tsai2019transformer}
Y.-H.~H. Tsai, S.~Bai, M.~Yamada, L.-P. Morency, and R.~Salakhutdinov.
\newblock Transformer dissection: An unified understanding for transformer’s
  attention via the lens of kernel.
\newblock In {\em Proceedings of the 2019 Conference on Empirical Methods in
  Natural Language Processing and the 9th International Joint Conference on
  Natural Language Processing (EMNLP-IJCNLP)}, pages 4335--4344, 2019.

\bibitem{transformer}
A.~Vaswani, N.~Shazeer, N.~Parmar, J.~Uszkoreit, L.~Jones, A.~N. Gomez, L.~u.
  Kaiser, and I.~Polosukhin.
\newblock Attention is all you need.
\newblock In I.~Guyon, U.~V. Luxburg, S.~Bengio, H.~Wallach, R.~Fergus,
  S.~Vishwanathan, and R.~Garnett, editors, {\em Advances in Neural Information
  Processing Systems 30}, pages 5998--6008. Curran Associates, Inc., 2017.

\bibitem{gran}
P.~Velickovic, G.~Cucurull, A.~Casanova, A.~Romero, P.~Li{\`{o}}, and
  Y.~Bengio.
\newblock Graph attention networks.
\newblock In {\em 6th International Conference on Learning Representations,
  {ICLR} 2018, Vancouver, BC, Canada, April 30 - May 3, 2018, Conference Track
  Proceedings}. OpenReview.net, 2018.

\bibitem{vig2019multiscale}
J.~Vig.
\newblock A multiscale visualization of attention in the transformer model.
\newblock {\em arXiv preprint arXiv:1906.05714}, 2019.

\bibitem{analyzing_attention}
J.~Vig and Y.~Belinkov.
\newblock Analyzing the structure of attention in a transformer language model.
\newblock {\em CoRR}, abs/1906.04284, 2019.

\bibitem{bertology}
J.~Vig, A.~Madani, L.~R. Varshney, C.~Xiong, R.~Socher, and N.~F. Rajani.
\newblock Bertology meets biology: Interpreting attention in protein language
  models.
\newblock {\em CoRR}, abs/2006.15222, 2020.

\bibitem{pointer}
O.~Vinyals, M.~Fortunato, and N.~Jaitly.
\newblock Pointer networks.
\newblock In {\em Advances in Neural Information Processing Systems 28: Annual
  Conference on Neural Information Processing Systems 2015, December 7-12,
  2015, Montreal, Quebec, Canada}, pages 2692--2700, 2015.

\bibitem{weigt2009identification}
M.~Weigt, R.~A. White, H.~Szurmant, J.~A. Hoch, and T.~Hwa.
\newblock Identification of direct residue contacts in protein--protein
  interaction by message passing.
\newblock {\em Proceedings of the National Academy of Sciences}, 106(1):67--72,
  2009.

\bibitem{shared_weights}
T.~Xiao, Y.~Li, J.~Zhu, Z.~Yu, and T.~Liu.
\newblock Sharing attention weights for fast transformer.
\newblock In S.~Kraus, editor, {\em Proceedings of the Twenty-Eighth
  International Joint Conference on Artificial Intelligence, {IJCAI} 2019,
  Macao, China, August 10-16, 2019}, pages 5292--5298. ijcai.org, 2019.

\bibitem{hans}
Z.~Yang, D.~Yang, C.~Dyer, X.~He, A.~J. Smola, and E.~H. Hovy.
\newblock Hierarchical attention networks for document classification.
\newblock In K.~Knight, A.~Nenkova, and O.~Rambow, editors, {\em {NAACL} {HLT}
  2016, The 2016 Conference of the North American Chapter of the Association
  for Computational Linguistics: Human Language Technologies, San Diego
  California, USA, June 12-17, 2016}, pages 1480--1489. The Association for
  Computational Linguistics, 2016.

\bibitem{energy}
H.~You, C.~Li, P.~Xu, Y.~Fu, Y.~Wang, X.~Chen, R.~G. Baraniuk, Z.~Wang, and
  Y.~Lin.
\newblock Drawing early-bird tickets: Toward more efficient training of deep
  networks.
\newblock In {\em International Conference on Learning Representations}, 2020.

\bibitem{ort}
F.~X. Yu, A.~T. Suresh, K.~M. Choromanski, D.~N. Holtmann{-}Rice, and S.~Kumar.
\newblock Orthogonal random features.
\newblock In D.~D. Lee, M.~Sugiyama, U.~von Luxburg, I.~Guyon, and R.~Garnett,
  editors, {\em Advances in Neural Information Processing Systems 29: Annual
  Conference on Neural Information Processing Systems 2016, December 5-10,
  2016, Barcelona, Spain}, pages 1975--1983, 2016.

\bibitem{relational}
V.~F. Zambaldi, D.~Raposo, A.~Santoro, V.~Bapst, Y.~Li, I.~Babuschkin,
  K.~Tuyls, D.~P. Reichert, T.~P. Lillicrap, E.~Lockhart, M.~Shanahan,
  V.~Langston, R.~Pascanu, M.~Botvinick, O.~Vinyals, and P.~W. Battaglia.
\newblock Deep reinforcement learning with relational inductive biases.
\newblock In {\em 7th International Conference on Learning Representations,
  {ICLR} 2019, New Orleans, LA, USA, May 6-9, 2019}, 2019.

\end{thebibliography}
